\documentclass[onecolumn]{article}

\usepackage[final]{corl_2021} % Uncomment for the camera-ready ``final'' version.
\usepackage{graphicx}
\usepackage{comment}
\usepackage{amsmath,amssymb} % define this before the line numbering.
\usepackage{multirow}
\usepackage{multicol}
\usepackage{subfigure}
\usepackage{algorithm} 
\usepackage{bm}
\usepackage{booktabs}
\usepackage{microtype}
\usepackage{arydshln}

\usepackage[misc]{ifsym} % to use \Letter
 
\title{Probabilistic and Geometric Depth: Detecting Objects in Perspective}

\author{
  Tai Wang$^{1,2}$ \quad Xinge Zhu$^1$ \quad Jiangmiao Pang$^{1,2}$\thanks{Corresponding author} \quad Dahua Lin$^{1,2,3}$ \\
  {\small $^1$CUHK-SenseTime Joint Lab, The Chinese University of Hong Kong}\\
  $^2$Shanghai AI Laboratory \quad $^3$Centre of Perceptual and Interactive Intelligence\\
  {\tt\small \{wt019, zx018, dhlin\}@ie.cuhk.edu.hk, pangjiangmiao@gmail.com} \\
}

\begin{document}

\maketitle

%===============================================================================

% !TEX root = ./arxiv.tex

\begin{abstract}
%----From Jiangmiao
3D object detection is an important capability needed in various practical applications such as driver assistance systems. 
Monocular 3D detection, as a representative general setting among image-based approaches, provides a more economical solution than conventional settings relying on LiDARs but still yields unsatisfactory results.
% Monocular 3D detection, as an economical solution compared to conventional settings relying on binocular vision or LiDAR, has drawn increasing attention recently but still yields unsatisfactory results. 
This paper first presents a systematic study on this problem. We observe that the current monocular 3D detection can be simplified as an instance depth estimation problem: The inaccurate instance depth \emph{blocks} all the other 3D attribute predictions from improving the overall detection performance.
Moreover, recent methods directly estimate the depth based on isolated instances or pixels while ignoring the geometric relations across different objects. 
% These geometric relations can be valuable constraints as the key information about depth is not directly manifest in the monocular image. 
To this end, we construct \emph{geometric} relation graphs across predicted objects and use the graph to facilitate depth estimation. As the preliminary depth estimation of each instance is usually inaccurate in this ill-posed setting, we incorporate a \emph{probabilistic} representation to capture the uncertainty. It provides an important indicator to identify confident predictions and further guide the depth propagation. Despite the simplicity of the basic idea, our method, PGD, obtains significant improvements on KITTI and nuScenes benchmarks, achieving 1st place out of all monocular vision-only methods while still maintaining real-time efficiency. Code and models will be released at \footnotesize\url{ https://github.com/open-mmlab/mmdetection3d}.

\end{abstract}
\vspace{-2ex}
\keywords{Probabilistic and Geometric Depth, Monocular 3D Detection} 

% !TEX root = ./arxiv.tex
\vspace{-1.5ex}
\section{Introduction}
\vspace{-0.8ex}
\label{sec:introduction}
% Merge original first two paragraphs: Background (importance of mono vs. 2d/LiDAR) + set up challenge (need depth but do not have depth)

3D object detection is an essential task for many robotic systems such as autonomous vehicles. Recent advanced methods in this field typically resort to various sensors, such as LiDAR~\cite{VoxelNet,PointPillars,STD,reconfig_voxels,ssn,Cylinder3D}, Radar~\cite{CenterFusion}, binocular vision~\cite{StereoRCNN,ida3d}, or their combinations for accurate depth information. Nevertheless, these perceptual systems are complicated, expensive, and difficult to maintain in complex environments. 
In contrast, monocular 3D detection, a setting that aims at perceiveing 3D objects from 2D monocular images, has drawn increasing attention due to its low costs.
However, as the depth information is not directly manifest in the input, this task is inherently ill-posed, making the problem particularly challenging.  

\vspace{-0.5ex}
This paper starts from a systematic study about this problem on two authoritative benchmarks in a quantitative way. Although we already knew the depth information is critical to this task, the study surprisingly shows that inaccurate depth estimation blocks all the other localization predictions from improving the final results. As instance depth has shown to be the bottleneck, we can simplify monocular 3D detection as an instance depth estimation problem to tackle it essentially.  

% However, recent monocular 3D object detection methods either use an extra network~\cite{MLFusion,PseudoLiDAR} for depth estimation or take depth as an ordinary component of 3D bounding boxes to optimize it~\cite{MonoDIS,M3D-RPN,RTM3D}.
% They commonly regard depth as one of the 3D attributes, predict it based on isolated instances or pixels and optimize it in a regression manner.
\vspace{-0.5ex}
Previous methods~\cite{MLFusion,PseudoLiDAR} first use an extra cumbersome depth estimation model to complement 2D detectors on depth information. 
The following methods~\cite{MonoDIS,M3D-RPN,RTM3D} simplify the frameworks by directly regarding depth as one dimension of the 3D localization task.
However, they still use simple methods that estimate depth from isolated instances or pixels in a regression manner.
We observe that aside from each object itself, other objects are co-existing in an image and the geometric relations across them can be valuable constraints to guarantee accurate estimation.

\begin{figure}
\begin{center}
\includegraphics[width=0.97\textwidth]{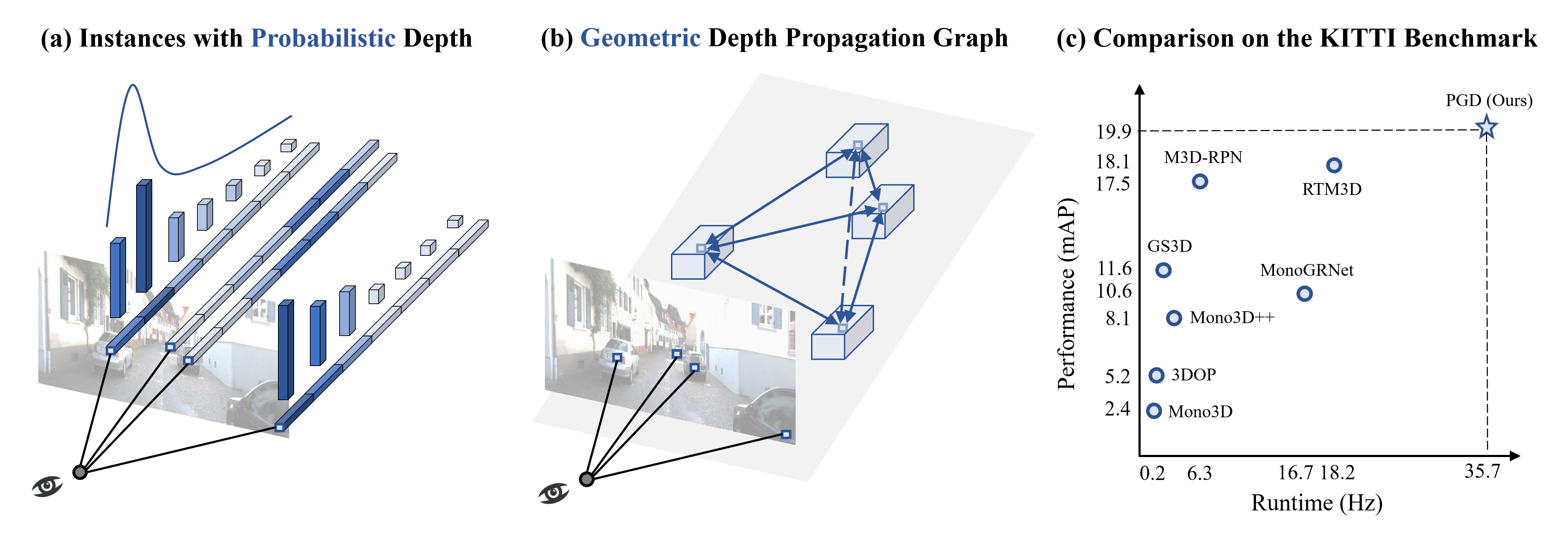}
\end{center}
   \vspace{-3.0ex}
   \caption{In this paper, to tackle the dominating depth estimation problem in the monocular 3D detection, we first (a) predict the depth of each instance with a probabilistic representation to capture the uncertainty, and (b) further construct a geometric relation graph to enhance the estimation from contextual connections. (c) The proposed method outperforms the other work in terms of both performance and speed significantly on the KITTI 3D car detection benchmark.}
\label{fig:teaser}
\vspace{-3.5ex}
\end{figure}

\vspace{-0.5ex}
Motivated by these observations, we propose Probabilistic and Geometric Depth (PGD) that jointly leverages probabilistic depth uncertainty and geometric relationships across co-existed objects for accurate depth estimation.
% Specifically, to measure the uncertainty of the estimated depth, we first bucket the depth interval into a set of discrete values and represent the depth by expectation of the distribution (Fig.~\ref{fig:teaser}(a)). The average of top-k confidence scores from the distribution is taken as a measure for uncertainty. With this measure, we further construct an uncertainty-aware, geometry-driven depth propagation graph to connect and enhance the instance estimations with their contextual relationship (Fig.~\ref{fig:teaser}(b)). Benefiting from this overall scheme, we can easily identify the predictions with higher confidence, and more importantly, estimate their depths more accurately with the graph-based synergistic mechanism.
Specifically, as the preliminary depth estimation of each instance is usually inaccurate in this ill-posed setting, we incorporate a probabilistic representation to capture the uncertainty of the estimated depth. We first bucket the depth values into a set of intervals and calculate the depth by the expectation of the distribution (Fig.~\ref{fig:teaser}(a)). The average of top-k confidence scores from the distribution is taken as the uncertainty of the depth.
To model the geometric relations, we further construct a depth propagation graph to enhance the estimations with their contextual relationship. The uncertainty of each instance depth provides useful guidance for the propagation therein.
Benefiting from this overall scheme, we can easily identify the predictions with higher confidence, and more importantly, estimate their depths more accurately with the graph-based synergistic mechanism.

% Evaluation results
% We evaluated our method on two 3D detection benchmarks with different settings and metrics, KITTI~\cite{KITTI} and nuScenes~\cite{nuScenes}. Based on a simple adapted monocular 3D detector FCOS3D~\cite{FCOS3D}, our fast real-time solution achieved the state-of-the-art on both KITTI and nuScenes, proving that with only designs tailored to depth, a detector could easily have the capability of detect objects in perspective.
%\zxg{Fig.1 is not involved; Maybe a teaser about instance depth estimation is better.}
\vspace{-0.6ex}
We implement the methods on a simple monocular 3D object detector FCOS3D~\cite{FCOS3D}.
Despite the simplicity of the basic idea, our PGD results in significant improvements on KITTI~\cite{KITTI} and nuScenes~\cite{nuScenes} with different benchmark settings and evaluation metrics. 
It achieves 1st place out of all monocular vision-only methods while still maintaining real-time efficiency.
The simple yet effective method proves that with only designs tailored to depth, a 2D detector can be capable of detecting objects in perspective.

% !TEX root = ./arxiv.tex

\vspace{-0.3ex}
\section{Related Work}
\vspace{-0.6ex}
%\jiangmiao{I recommend to name the Sec. 2.1 ``2D Object Detection", as our method can be generally extended many detectors.}\\
%\jiangmiao{There's some probabilistic scene depth estimation methods. As there is a Sec. 2.2, we need to discuss the difference vs. them.}\\
%\zxg{Comparison between global depth estimation and instance depth; Show some work focusing on global-scene depth and illustrate its drawback.}
%\jiangmiao{We can separate Sec. 2.3 into several paragraphs, use \#noindent + textbf\# to categorize and highlight them.}
%\jiangmiao{This part is too long that occupy too much space. Generally I think half or 3/4 page is good for related work. We can simplify Sec.2.1 as it is only the background not the problem we are solving, 1-2 paragraphs is good for it. Just try to summarize some papers and cite them together, only talk the papers in depth if they are indeed similar/related to our work. Btw, use paragraph hightlight to inplace the subsection, and emph for the exisiting paragraph highlight also help saving space.}
%\TODO{One para for 2D det, simplify the 2.2 and use textbf to replace the subsections.}

\noindent\textbf{2D Object Detection}\quad
According to the base of initial guesses, modern 2D detection methods can be divided into two branches, anchor-based and anchor-free. Anchor-based methods~\cite{FastRCNN,FasterRCNN,SSD,YOLOv2} benefit from the predefined anchors in terms of much easier regression, while anchor-free methods~\cite{DenseBox,YOLOv1,FCOS,CornerNet,CenterNet} do not need complicated prior settings and thus have better universality. For simplicity, we take FCOS3D~\cite{FCOS3D}, the 3D adapted version of FCOS~\cite{FCOS}, as the baseline considering its capability of handling overlapped ground truths and scale variance problem.

\vspace{-0.3ex}
\noindent\textbf{Monocular Depth Estimation}\quad
Monocular depth estimation is also a challenging ill-posed problem like monocular 3D detection. It aims at predicting dense and global depth field at pixel level given an RGB image. Early works~\cite{LearnDepth,Make3D} predict depth from hand-crafted features with non-parametric optimization methods. With the rapid progress of CNNs, fully supervised methods~\cite{Eigen,supervised1,supervised2}, self-supervised methods based on stereo pairs~\cite{stereo1,stereo2} and monocular videos~\cite{video1,video2} gradually emerged.
Although this problem has been explored for a long time, there are very few works~\cite{Foresee} studying it in a specific task, like detection, where the dense depth supervision is always not guaranteed and we only care about the accuracy of instance depth instead of the global depth field.

\vspace{-0.6ex}
As for the reformulation of depth learning problems, there are a few attempts in this field.
For example, DORN~\cite{DORN} recasts the depth learning problem as ordinal regression and proposes a spacing-increasing discretization (SID) strategy to improve network training and reduce computations. It is similar to the underlying idea of our probabilistic representation for uncertainty modeling while different in terms of motivation and design details.

\vspace{-0.3ex}
\noindent\textbf{Monocular 3D Object Detection}\quad
Monocular 3D detection is more complicated than the 2D case. The underlying problem is the inconsistency of input 2D data modal and the output 3D predictions.

\vspace{-0.6ex}
\noindent\emph{Methods involving sub-networks}\quad Earlier work uses sub-networks to assist 3D detection. 3DOP~\cite{3DOP} and MLFusion~\cite{MLFusion} use a depth estimation network while Deep3DBox~\cite{Deep3DBox} uses a 2D object detector. They rely on the design and performance of these sub-networks, even external data and pre-trained models, which makes the training inconvenient and introduces additional system complexity.
%implicitly sets an upper bound to the overall performance.
%\jiangmiao{Why `implicitly sets an upper bound to the overall performance`}

\vspace{-0.6ex}
\noindent\emph{Transform to 3D representation}\quad Another category is to convert the RGB input to 3D representations like OFTNet~\cite{OFTNet} and Pseudo-Lidar~\cite{PseudoLiDAR}. Although these methods have shown promising performance, they actually rely on dense depth labels and hence are not regarded as pure monocular approaches.
There are also domain gaps between different depth sensors, making them hard to generalize smoothly to a new practical setting. Furthermore, the efficiency of processing a large number of point clouds is also a significant issue to deal with in practical applications.

\vspace{-0.6ex}
\noindent\emph{End-to-end designs like 2D detection}\quad Recent work notices these drawbacks, and end-to-end frameworks are thus proposed. M3D-RPN~\cite{M3D-RPN} implements a single-stage multi-class detector with an end-to-end region proposal network and depth-aware convolution. SS3D~\cite{SS3D} proposes to detect 2D key points and further predicts object characteristics with uncertainties. MonoDIS~\cite{MonoDIS} introduces a disentangling loss to reduce the instability of the training procedure. Some of them still have multiple training stages or post-optimization phases. In addition, they all follow anchor-based manners, and thus the consistency of 2D and 3D anchors is needed to be determined. In contrast, anchor-free methods~\cite{CenterNet,RTM3D,MonoPair,FCOS3D} do not need to make statistics on the given data and have better generalized ability to more various classes or different intrinsic settings, so we choose to follow this paradigm.
%\TODO{just say the difference between anchor-based and anchor-free monocular detectors, and we follow anchor-free.}

Nevertheless, all of these works rarely have customized designs for instance depth estimation in particular, and only take it as one common regression target for isolated points or instances. It actually hinders the breakthrough of this problem, which will be discussed in our quantitative study and specifically addressed in our approach.
\label{sec:related}

% !TEX root = ./arxiv.tex
\begin{figure}
\begin{center}
\includegraphics[width=1.0\linewidth]{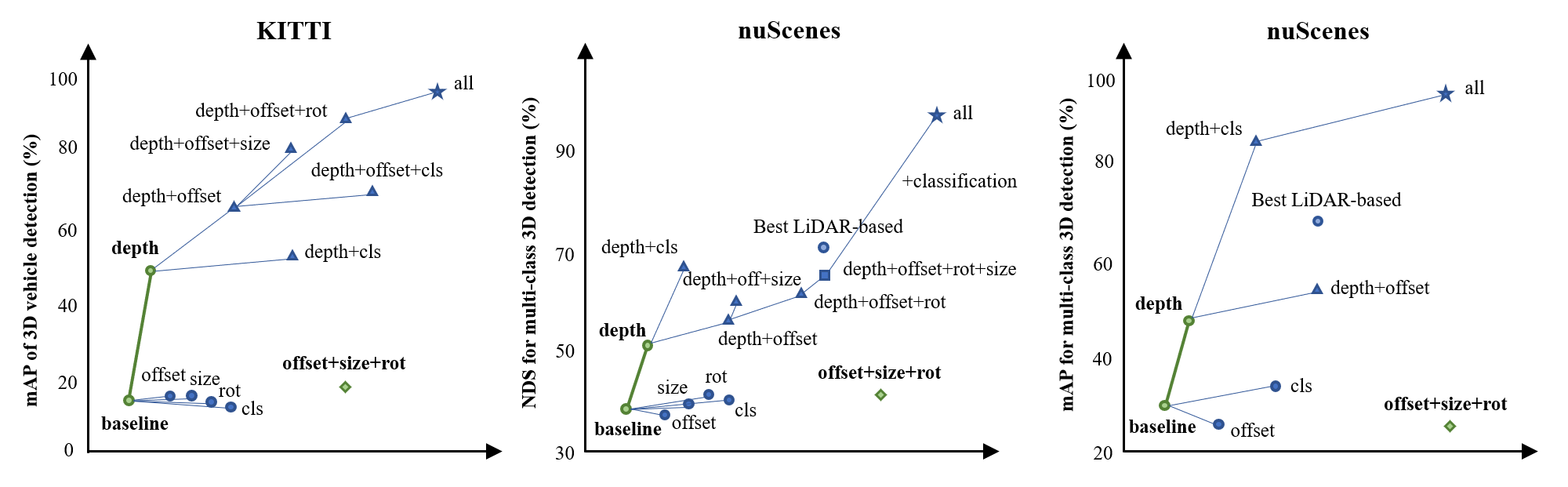}
\end{center}
   \vspace{-3.5ex}
   \caption{Oracle analyses with different datasets and metrics. From left to right: 3D IoU based mAP on KITTI, NuScenes Detection Score (NDS) and distance-based mAP on nuScenes. We replace our predictions with ground truth values step by step and observe the performance improvements. It can be seen that an accurate depth can bring significant performance improvement (green lines), and only with accurate depth can the improvements brought by other oracles be realized.}
\label{fig: oracle}
\vspace{-3.5ex}
\end{figure}

\vspace{-0.8ex}
\section{Preliminary and Motivating Study}
\vspace{-0.8ex}
In this section, we aim at making an in-depth quantitative error analysis on top of a basic adapted monocular 3D detector to investigate the key challenge in the specific 3D detection setting.

\begin{figure}
\begin{center}
\includegraphics[width=1.0\linewidth]{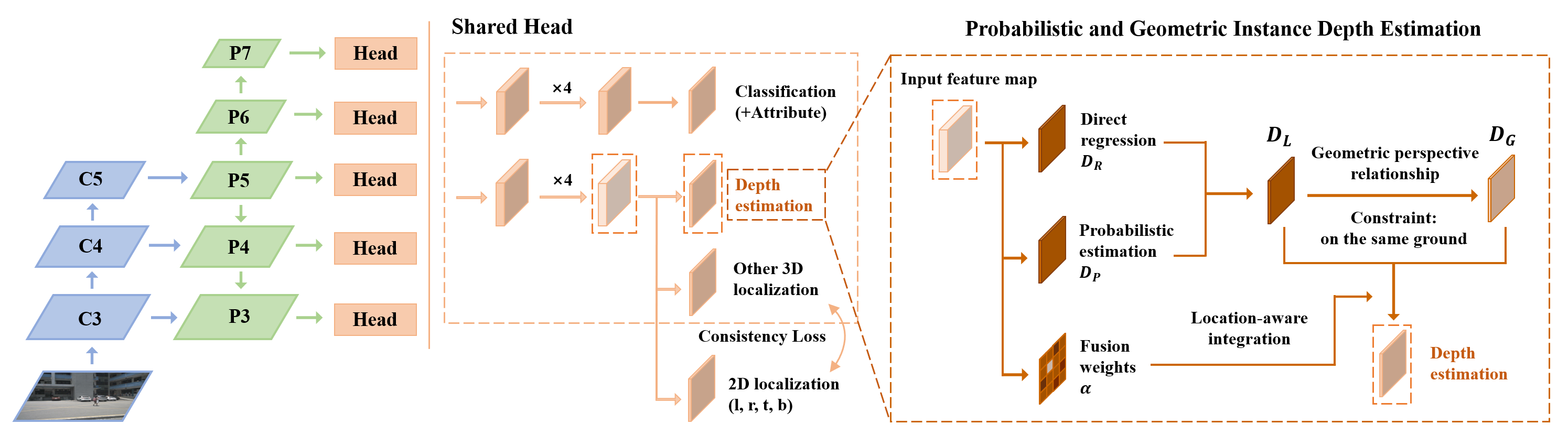}
\end{center}
   \vspace{-3.0ex}
   \caption{An overview of our framework. We start from a basic monocular 3D detector, FCOS3D, while focus on tackling the difficulty of instance depth estimation with our proposed customized module in the head. With the feature map from the regression branch as the input, we first introduce a branch for probabilistic depth estimation to model the uncertainty, then derive the geometric depth with the depth propagation graph and finally integrate them to get the final depth prediction.}
   %\TODO{Use black as the basic color. Simplify the outputs of shared head. Make the font size seem similar with that in the main paper.}
\label{fig: overview}
\vspace{-3ex}
\end{figure}

Typically, conventional 2D detection expects the model to predict 2D bounding boxes and category labels for each object of interest, while a monocular 3D detector needs to predict 7-DoF 3D boxes given the same input. From this perspective of problem formulation, the main difference lies on the regression targets. An intuitive reason for the much worse performance of monocular 3D detection compared to 2D is that there exist much more difficult targets to regress in the \emph{localization}. Hence, we choose a simple detector FCOS3D~\cite{FCOS3D} to study the specific problem, which keeps the well-developed designs for 2D feature extraction and is adapted for this 3D task with only basic designs for specific 3D detection targets. As shown in the left part of Fig.~\ref{fig: overview}, there are overall two branches for classification and localization respectively. Formally, for the regression branch, the detector predicts 3D attributes, including offsets $\Delta$x, $\Delta$y to the projected 3D center, depths $d$, 3D size $w^{3D}$, $l^{3D}$, $h^{3D}$, sin value of rotation $\theta$, direction class $C_{\theta}$, center-ness $c$, and distances to four sides of 2D boxes $l$, $r$, $t$, $b$, for each location on the output dense map. We further equip it with a basic consistency loss between 3D and 2D localization, which will be detailed in the appendix.

On this basis, we apply this baseline on two representative benchmarks, KITTI and nuScenes, and replace the predictions with ground truths step by step to identify the performance bottleneck (Fig.~\ref{fig: oracle}). Unexpectedly, the \emph{inaccurate depth} blocks \emph{all} the other sub-task predictions from improving the overall detection performance, on both datasets under different metrics. Hence, current monocular 3D detection, especially 3D localization, can be reduced to the dominating instance depth estimation problem to a great extent, which will be the focus of our method to be presented next. See more details about the explanation of oracle analyses in the appendix.

\vspace{-0.5ex}
\section{Our Approach}
\vspace{-0.5ex}

Given images collected from similar cameras, previous work typically resorts to direct regression for instance depth estimation and expects the model to directly learn that objects with certain appearances and sizes always exist at locations with certain depths. Our baseline also follows this way. However, it is hard to learn due to the large variance and also obviously not enough for the accuracy needed in 3D detection. Given the inherent downside of hard regression for isolated points, in our approach, we aim at constructing an uncertainty-aware depth propagation graph to enhance the estimation from contextual connections among instances. Next, we will first elaborate on the adopted probabilistic representation and technical details of the constructed geometric graph, and finally present how we integrate these obtained depth estimations.

\vspace{-0.5ex}
\subsection{Uncertainty Modeling with Probabilistic Representation}
\vspace{-0.5ex}
For a one-stage detector, a general design for direct depth estimation is a small head along the regression branch expected to output a dense depth map. Formally, as shown in Fig.~\ref{fig: overview}, suppose the input feature map has shape $H\times W$, then the direct depth regression output~\footnote{To make learning easier, the output of direct regression branch is applied an exponential transformation.} can be denoted as $D_R\in\mathbb{R}^{H\times W}$. On this basis, to establish an effective depth propagation mechanism, modeling the uncertainty of depth estimation for each instance is an important preliminary, which can provide useful guidance for weighing the propagation among instances. We adopt a simple yet effective probabilistic representation to achieve this: Considering the depth value is continuous in a certain range, we uniformly quantize the depth interval into a set of discrete values and represent the prediction with the expectation of the distribution.
Suppose the detection range is $0\sim D_{max}$, the discretized unit is $U$, then we have $C=\lfloor D_{max}/U\rfloor +1$ split points. Denote the set of points as a weight vector $\omega\in\mathbb{R}^{C}$, and then we introduce a new head parallel with direct regression to produce a probabilistic output map $D_{PM}$, which will be decoded with:
\vspace{-1.0ex}
\begin{small}
\begin{equation}
    D_P = \omega^T softmax(D_{PM})
\end{equation}
\end{small}
\hspace{-1.5ex} where $D_P$ is the so-called probabilistic depth. It is equivalent to compute the expectation of the probabilistic distribution formed by $softmax(D_{PM})$. Apart from the $D_P$, we can further obtain the depth confidence score, denoted as $s^d\in S_D$, from the depth distribution of each instance. In practice, we take the average of top-2 confidence as the depth score for $U=10m$. It will be multiplied by the center-ness and classification score as the final ranking criterion for predictions during inference.

Subsequently, we fuse $D_R$ and $D_P$ with the sigmoid response of a data-agnostic single parameter $\lambda$:
\vspace{-0.5ex}
\begin{small}
\begin{equation}
    D_L = \sigma(\lambda)D_R + (1 - \sigma(\lambda))D_P
\end{equation}
\end{small}
\hspace{-1.5ex} Here $D_L$ is regarded as a \emph{local} depth estimation for each isolated instance, which together with the depth score derived from $D_{PM}$ serve as the foundation of constructing the depth propagation graph.

It is worth noting that our implementation is different from the typical way used in monocular depth estimation~\cite{DORN}, which usually adopts a fine-grained quantization for the depth interval and further estimates the value with classification and residual regression. In comparison, our method is more memory-efficient, more straightforward for regressing continuous value, and provides a natural indicator for uncertainty estimation. Please refer to the appendix for empirical results about comparison with other depth interval division methods.

% We can benefit from this design mainly on two aspects. The first is it could complement the direct regression especially for long range, because the latter will be less sensitive to changes of large numbers due to the log transformation. The second is we can obtain the confidence score for depth prediction from the distribution, which can also be leveraged for inference. In practice, we take the average of top-2 confidence as the depth score $s_d$ for $U=10m$ and multiply it with classification score as well as center-ness as the final rank criterion, which brings a significant gain especially for strict metrics. Finally, we also empirically find that the optimization is smoother and more efficient with this representation. 

\begin{figure}
\begin{center}
\includegraphics[width=0.9\linewidth]{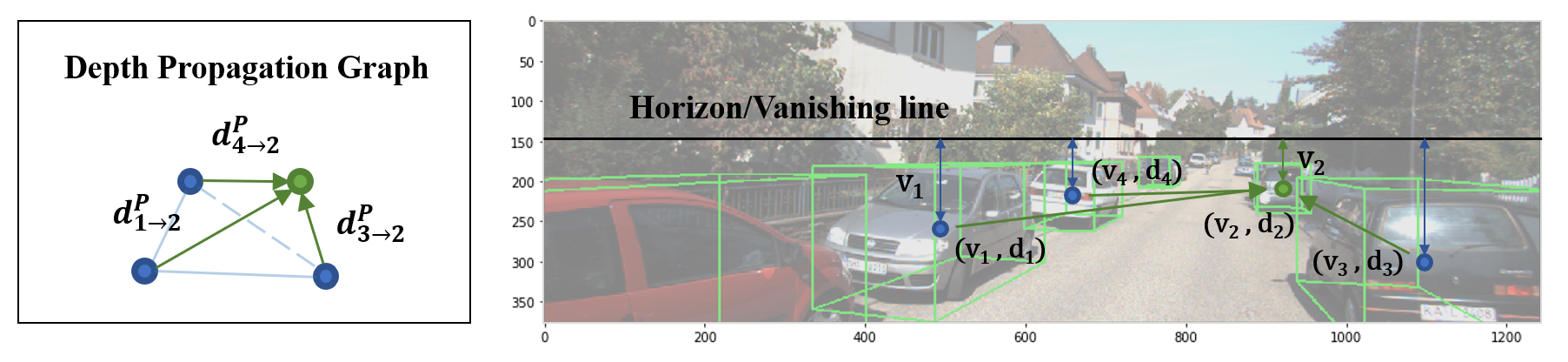}
\end{center}
   \vspace{-3.5ex}
   \caption{For objects hard to approximate depth accurately (like car 2), propagating (green arrows) reliable depth predictions from other objects (car 1, 3, 4) with perspective geometry can enhance the reasoning from the global context.}
\label{fig: depth_prop}
\vspace{-3.5ex}
\end{figure}

\vspace{-0.5ex}
\subsection{Depth Propagation from Perspective Geometry}\label{sec: depth_prop}
\vspace{-0.5ex}
With the depth prediction $D_L$ for isolated instances and their depth confidence scores $S_D$ for uncertainty estimation, we can further construct the propagation graph based on the contextual geometric relationship. Consider the typical driving scenarios: a general constraint can be leveraged, \emph{i.e.}, almost all the objects are on the ground. Early in~\cite{PutPesp}, Hoiem et al. utilized the scene projection with this constraint to put objects in the context of the overall 3D scene by modeling the relationship of different elements. Here, targeting the depth estimation problem, we instead propose a geometric depth propagation mechanism with consideration of interdependence between instances. Next, we will first derive the perspective relationship between two instances, and then present the details of the graph-based depth propagation scheme with edge pruning and gating.

\noindent\textbf{Perspective Relationship}\quad
Consider in the general perspective projection, suppose the camera projection matrix $P$ is:
\vspace{-1.0ex}
\begin{equation}
    P =  \left(
    \begin{matrix}
    f & 0 & c_u & -fb_x \\
    0 & f & c_v & -fb_y \\
    0 & 0 & 1 & -fb_z
    \end{matrix}
    \right)
    \vspace{-0.4ex}
\end{equation}
where $f$ is the focal length, $c_u$ and $c_v$ are the vertical and horizon position of camera in the image, $b_x$, $b_y$ and $b_z$ denote the baseline with respect to the reference camera (non-zero in KITTI while zero in nuScenes). Note that we represent the focal length with a single $f$ considering most cameras share the same one for the $u$ and $v$ axis. Then a 3D point $\mathbf{x^{3D}}=(x,y,z,1)^T$ in the camera coordinates can be projected to a point $\mathbf{x^{2D}}=(u',v',1)^T$ in the image with:
\vspace{-1.0ex}
\begin{equation}
d\mathbf{x_{2D}}=P\mathbf{x_{3D}}
\end{equation}
To simplify the result, we replace $v'$ with $v+c_v$, then $v$ represents the distance to the horizon line (Down is the positive direction in Fig.~\ref{fig: depth_prop}). Then we get:
\vspace{-1.0ex}
\begin{equation}
    vd = f(y-b_y+c_v b_z)
    \label{eqn: proj}
    \vspace{-0.4ex}
\end{equation}
The relation for $u$ is similar. Considering the constraint that all the objects are on the ground, the bottom centers of objects always share the same $y$ (height in the camera coordinates), so we mainly consider this relation for $v$ next. Given two objects 1 and 2, the relationship between their center depths can be derived from Eqn.~\ref{eqn: proj}:
\vspace{-1.0ex}
\begin{equation}
    d_2 = \frac{v_1}{v_2}d_1 + \frac{f}{v_2}(y_2-y_1) \approx \frac{v_1}{v_2}d_1 + \frac{f}{2v_2}(h^{3D}_1-h^{3D}_2) \triangleq d_{1\to2}^P
    \vspace{-0.4ex}
\end{equation}
with which we can predict $d_2$ given $d_1$ precisely with the height difference between 3D centers. Besides, we can also leverage an approximation of this relationship given the assumption that objects share the same bottom height, then $y_2-y_1$ can be substituted by the difference of half heights of 3D boxes $\frac{1}{2}(h^{3D}_1-h^{3D}_2)$, defined as $d_{1\to2}^P$.

In this relation, when $h^{3D}_1=h^{3D}_2$, $v_1d_1=v_2d_2$, which is easy to understand, \emph{i.e.}, an object closer to vanishing line is farther away. It is a clear relationship connecting different instances but also yields errors. Suppose $|(y_2-y_1)-\frac{1}{2}(h_1-h_2)|=\delta$, the error of depth will be $\Delta d=\frac{f}{v_2}\delta$. When $\delta=0.1m$, $v_2=50$ (pixels), $\Delta d$ can be about $1.5m$. Although it is acceptable for objects $30m$ away (corresponding with $v_2=50$), we also need a mechanism to avoid possible large errors. It consists of the edge pruning and gating scheme to be described next and the location-aware weight map to be mentioned in the Sec.~\ref{sec: integration}.

\noindent\textbf{Graph-Based Depth Propagation}\quad
With the pairwise perspective relationship, we can estimate the depth of any object from the cues of other objects. Then we can construct a dense directed graph with two bidirectional edges between any two objects representing the depth propagation (Fig.~\ref{fig: depth_prop}). Formally, suppose we have $N$ predicted objects with indices from $\mathcal{P}=\{1,2,...,n\}$, we can estimate the depth of object $i$ given $d_{j\to i}^P$ for all the $j\in \mathcal{P}$, defined as the geometric depth $d_i^G\in D_G$. Considering the computational efficiency and possible large errors mentioned previously, we propose an edge pruning and gating scheme to improve the propagation graph. From our observation, the same category of nearby objects can well satisfy the "same ground" condition, so we select the following 3 most important factors to decide which edges are influential and reliable, including the depth confidence $s^d_j$, 2D distance score $s^{2D}_{ij}$, and classification similarity $s^{cls}_{ij}$. The latter two and the overall edge score $s^e_{j\to i}$ are computed as follows:
\vspace{-1.0ex}
\begin{equation}
    s^{2D}_{ij} = 1-\frac{t^{2D}_{ij}}{t^{2D}_{max}},\quad s^{cls}_{ij} = \frac{\bm{f_i}\cdot \bm{f_j}}{||\bm{f_i}||_2||\bm{f_j}||_2},\quad s^e_{j\to i} = \frac{s^d_j\cdot s^{2D}_{ij}\cdot s^{cls}_{ij}}{\sum_{j=1}^k s^d_j\cdot s^{2D}_{ij}\cdot s^{cls}_{ij}}
\end{equation}
where $t^{2D}_{ij}$ is the 2D distance between projected centers of object $i$ and $j$, $t^{2D}_{max}$ is set to the length of image diagonal, $\bm{f_i}$ and $\bm{f_j}$ are the output confidence vectors of two objects from classification branch and $k$ is the maximum number of edges to be kept after pruning (edges with top-k scores are kept). The edge score is then used for gating so that each node attends its edges with their importance:
\vspace{-1.0ex}
\begin{equation}
    d_i^G = \sum_{j=1}^{k}s^e_{j\to i}d_{j\to i}^P
\end{equation}
Note that obtaining the geometric depth map $D_G$ from this graph is free of learnable parameters. To avoid influencing the learning of other components, we cut off the gradients backpropagated from this computation and only focus on how to integrate $D_L$ and $D_G$, which will be discussed next.

\subsection{Probabilistic and Geometric Depth Estimation}\label{sec: integration}
So far, we have obtained two depth predictions $D_L$ and $D_G$ from isolated and graph-based contextual estimations, respectively. Then we integrate these two complementary components in a learning manner. Unlike the data-agnostic single parameter used in the local estimation, integrating these two results should be more complex considering their flexible roles in various complicated cases. So we further introduce a branch to produce a location-aware weight map $\alpha\in\mathbb{R}^{H\times W}$ to fuse them (Fig.~\ref{fig: overview}):
\vspace{-0.4ex}
\begin{small}
\begin{equation}
    D = \sigma(\alpha)\circ D_L + (1 - \sigma(\alpha))\circ D_G
    \label{eqn: global_fuse}
\end{equation}
\end{small}
\hspace{-1.0ex} The fused depth $D$ will replace the direct regressed $D_R$ in the baseline and trained with the common smooth L1 loss in the same end-to-end way. Note that adding intermediate supervisions empirically makes the training more stable but does not bring any performance gains.
% !TEX root = ./arxiv.tex
\vspace{-0.5ex}
\section{Experiments}\label{sec:experimental_setup}
\vspace{-0.5ex}
In this section, we present our experimental setup and implementation details, and then make the quantitative analysis on the KITTI and nuScenes dataset with details of both performance and efficiency. Finally, detailed ablation studies are conducted to show the efficacy of each component in our method. Refer to the appendix for more qualitative analysis.
\subsection{Datasets \& Evaluation Metrics}
We evaluate our method on two datasets, KITTI~\cite{KITTI} and nuScenes~\cite{nuScenes}. There are 7481/7518 samples for training/testing respectively on KITTI, and the training samples are generally divided into 3712/3769 samples as training/validation splits. We first validate our method on this popular benchmark. Nevertheless, the variety of scenes and categories is limited on KITTI, so we further test our approach on the large-scale nuScenes dataset. NuScenes consists of multi-modal data collected from 1000 scenes, including RGB images from 6 cameras, points from 5 Radars, and 1 LiDAR. It is split into 700/150/150 scenes for training/validation/testing. There are overall 1.4M annotated 3D bounding boxes from 10 categories. In addition, nuScenes uses different metrics, distance-based mAP and NDS, which can help evaluate our method from another perspective. See more explanations about metrics in the appendix.
\subsection{Implementation Details}
\noindent\textbf{Network Architectures}\quad As shown in Fig.~\ref{fig: overview}, our baseline framework basically follows the design of FCOS3D~\cite{FCOS3D}. Given the input image, we utilize ResNet101~\cite{ResNet} as the feature extraction backbone followed by FPN~\cite{FPN} for generating multi-level predictions. Detection heads are shared among multi-level feature maps except that three scale factors are used to differentiate some of their final regressed results, including offsets, depths, and sizes, respectively. For the hyperparameters in the depth estimation module, $U$ is set to $10m$ and $k$ is set to 5. The overall framework is built on top of MMDetection3D~\cite{mmdet3d2020}. Please refer to FCOS3D~\cite{FCOS3D} and appendix for the design of loss and other implementation details.

\noindent\textbf{Training Parameters}\quad For all the experiments, we trained randomly initialized networks from scratch following end-to-end manners. Models are trained with SGD optimizer, in which gradient clip and warm-up policy are exploited with learning rate 0.001, number of warm-up iterations 500, warm-up ratio 0.33 and batch size 32/12 on 16/4 GTX 1080Ti GPUs for nuScenes/KITTI.

\noindent\textbf{Data Augmentation}\quad We only implement image flip for augmentation, where offset and 2D targets are flipped for the 2D image while 3D boxes are transformed correspondingly in 3D space. No other augmentation (right image augmentation, cropping, resizing, \emph{etc.}) methods are utilized.

\begin{table}
\scriptsize
\caption{Results on the KITTI validation dataset}
\vspace{-2.5ex}
\begin{center}
\begin{tabular}{c|c|c|c|c|c|c|c|c|c}
\hline
 \multirow{2}*{Methods} & \multirow{2}*{Venue} &
 \multirow{2}*{Extra Labels} & \multirow{2}*{Time} & \multicolumn{3}{c|}{AP$_{BEV}$ IOU$\ge 0.7$} & \multicolumn{3}{c}{AP$_{3D}$ IOU$\ge 0.7$}\\
\cline{5-10}
~ & ~ & ~ & ~ & Easy & Mod. & Hard & Easy & Mod. & Hard\\
\hline
Mono3D~\cite{Mono3D} & CVPR 2016 & Mask & 4.2s & 5.22 & 5.19 & 4.13 & 2.53 & 2.31 & 2.31\\
3DOP~\cite{3DOP} & TPAMI 2017 & Stereo & 3s & 12.63 & 9.49 & 7.59 & 6.55 & 5.07 & 4.10\\
MF3D~\cite{MLFusion} & CVPR 2018 & Dense Depth & - & 22.03 & 13.63 & 11.60 & 10.53 & 5.69 & 5.39\\ 		
Mono3D++~\cite{Mono3D++} & AAAI 2018 & Dense Depth+Shape & $>$0.6s & 16.70 & 11.50 & 10.10 & 10.60 & 7.90 & 5.70\\
PL~\cite{PseudoLiDAR,Foresee} (AVOD) & CVPR 2019 & Dense Depth & - & 19.0 & 15.3 & 13.0 & 7.5 & 6.1 & 5.4\\
ForeSeE~\cite{Foresee} (AVOD) & AAAI 2020 & Dense Depth & - & 23.4 & 17.4 & 15.9 & 15.0 & 12.5 & 12.0\\
\hline
Deep3DBox~\cite{Deep3DBox} & CVPR 2018 & None & - & 9.99 & 7.71 & 5.30 & 5.85 & 4.10 & 3.84\\
MonoGRNet~\cite{MonoGRNet} & AAAI 2019 & None & 0.06s & - & - & - & 13.88 & 10.19 & 7.62\\
FQNet~\cite{FQNet} & CVPR 2019 & None & 3.33s & 9.50 & 8.02 & 7.71 & 5.98 & 5.50 & 4.75\\
GS3D~\cite{GS3D} & CVPR 2019 & None & 2.3s & - & - & - & 13.46 & 10.97 & 10.38\\
M3D-RPN~\cite{M3D-RPN} & ICCV 2019 & None & 0.16s & 25.94 & 21.18 & 17.90 & 20.27 & 17.06 & 15.21\\
MonoDIS~\cite{MonoDIS} & ICCV 2019 & None & - & 24.26 & 18.43 & 16.95 & 18.05 & 14.98 & 13.42 \\
RTM3D~\cite{RTM3D} & ECCV 2020 & None & 0.055s & 25.56 & 22.12 & \textbf{20.91} & 20.77 & 16.86 & 16.63 \\
\hline
FCOS3D~\cite{FCOS3D} & ICCVW 2021 & None & - & 18.16 & 14.02 & 13.85 & 13.90 & 11.61 & 10.98 \\
PGD (Ours) & - & None & \textbf{0.028s} & \textbf{30.56} & \textbf{23.67} & 20.84 & \textbf{24.35} & \textbf{18.34} & \textbf{16.90}\\
\hline
\end{tabular}
\end{center}
\label{tab: KITTI_val}
\vspace{-4.0ex}
\end{table}

\begin{table}\scriptsize
\caption{Results on the nuScenes dataset.}
\vspace{-2.5ex}
	\begin{center}
	\begin{tabular}{c|c|c|c|c|c|c|c|c|c}
	\hline
	Methods & Split & Modality & mAP & mATE & mASE & mAOE & mAVE & mAAE & NDS\\
	\hline
	PointPillars (Light) \cite{PointPillars} & test & LiDAR & 0.305 & 0.517 & 0.290 & 0.500 & 0.316 & 0.368 & 0.453\\
	CenterFusion~\cite{CenterFusion} & test & Cam. \& Radar & 0.326 & 0.631 & 0.261 & 0.516 & 0.614 & 0.115 & 0.449\\
	CenterPoint v2~\cite{CenterPoint} & test & Cam. \& LiDAR \& Radar & \textbf{0.671} & 0.249 & 0.236 & 0.350 & 0.250 & 0.136 & \textbf{0.714}\\
	\hline
	LRM0 & test & Camera & 0.294 & 0.752 & 0.265 & 0.603 & 1.582 & 0.14 & 0.371\\
	MonoDIS~\cite{MonoDIS} & test & Camera & 0.304 & 0.738 & 0.263 & 0.546 & 1.553 & 0.134 & 0.384\\
	CenterNet~\cite{CenterNet} & test & Camera & 0.338 & 0.658 & 0.255 & 0.629 & 1.629 & 0.142 & 0.4\\
	Noah CV Lab & test & Camera & 0.331 & 0.660 & 0.262 & 0.354 & 1.663 & 0.198 & 0.418\\
	FCOS3D~\cite{FCOS3D} & test & Camera & 0.358 & 0.690 & 0.249 & 0.452 & 1.434 & 0.124 & 0.428\\
	PGD (Ours) & test & Camera & \textbf{0.386} & 0.626 & 0.245 & 0.451 & 1.509 & 0.127 & \textbf{0.448}\\
	\hline
	CenterNet~\cite{CenterNet} & val & Camera & 0.306 & 0.716 & 0.264 & 0.609 & 1.426 & 0.658 & 0.328\\
	FCOS3D~\cite{FCOS3D} & val & Camera & 0.343 & 0.725 & 0.263 & 0.422 & 1.292 & 0.153 & 0.415\\
	PGD (Ours) & val & Camera & \textbf{0.369} & 0.683 & 0.260 & 0.439 & 1.268 & 0.185 & \textbf{0.428}\\
	\hline
	\end{tabular}
	\end{center}
	\label{tab: nus_quant}
	\vspace{-7.0ex}
\end{table}

\vspace{-0.5ex}
\subsection{Quantitative Analysis}
\vspace{-0.5ex}
We make quantitative analyses both on KITTI (Tab.~\ref{tab: KITTI_val} and \ref{tab: KITTI_test}, Fig.~\ref{fig:teaser}(c)) and much harder, less commonly validated nuScenes dataset (Tab.~\ref{tab: nus_quant}). It can be seen that our method achieves the state-of-the-art on both benchmarks with different settings and metrics while maintains outstanding speed.

We list part of early monocular methods with extra data or pre-trained models and recent image-only methods that have related results for comparison on the KITTI dataset. Only the results for car detection are compared here because the performance of small objects is always unstable due to their limited samples. Our framework based on the simple adapted FCOS3D achieves much better performance than others, especially considering M3D-RPN~\cite{M3D-RPN} and RTM3D~\cite{RTM3D} adopt stronger backbone and data augmentation. Furthermore, our method can run at the speed of 36Hz to achieve this, thanks to most of our modules not introducing extra computational costs to inference. It is an excellent trade-off between performance and efficiency.

Then for the nuScenes dataset, we also compare the results on the test set and validation set, respectively. On the test set, we first compared all the methods using RGB images as the input data. Our single model achieved the best performance among them with mAP 37.0\% and NDS 43.2\%, in which we particularly exceeded the previous best method more than 3\% in terms of mAP. We also list benchmarks based on other data modality, including lightweight, real-time PointPillars~\cite{PointPillars} with LiDAR, CenterFusion~\cite{CenterFusion} with RGB image and Radar, and CenterPoint~\cite{CenterPoint} ensemble results with all the sensors. It can be seen that although our method has a certain gap with the high-performance CenterPoint, it even surpasses PointPillars and CenterFusion on mAP, which shows that this ill-posed problem can be solved decently with enough data. At the same time, the methods using other modal data usually yield better NDS, mainly because the mAVE is smaller. The reason is that they can predict the speed of objects from continuous multi-frame point clouds or velocity measurement of Radar. In contrast, we only use the single-frame image in our experiments. So how to mine the speed information from consecutive frame images will be a direction worthy of exploring in the future. On the validation set, we compare our method with the best open-source center-based detector, CenterNet. Our method is not only much more efficient to train and inference (3 days to train the CenterNet vs. only one day to train our model with comparable performance), but also achieves better performance, especially in terms of the mAP and mAOE. On this basis, we finally achieved an improvement of about 9\% on NDS. See more detailed results about depth estimation accuracy and per-class detection performance in the appendix.

\begin{table}
\tiny
\begin{minipage}{.48\linewidth}
    \centering
    \caption{Ablation study on KITTI.}
    \vspace{-1.0ex}
    \label{tab: KITTI_ablation}
    \begin{tabular}{c|c|c|c|c|c|c}
    \hline
     \multirow{2}*{Method} & \multicolumn{3}{c|}{AP$_{3D}$ IOU$\ge 0.7$} & \multicolumn{3}{c}{AP$_{3D}$ IOU$\ge 0.5$}\\
    \cline{2-7}
    ~ & Easy & Mod. & Hard & Easy & Mod. & Hard\\
    \hline
    FCOS3D~\cite{FCOS3D} & 9.55 & 5.51 & 4.78 & 34.29 & 25.78 & 23.66\\
    +Local cons. & 14.62 & 12.42 & 11.02 & 39.11 & 26.86 & 25.62\\
    +Prob. depth & 19.10 & 16.04 & 14.83 & 47.64 & 37.45 & 33.29\\
    +Depth prop. & 21.36 & 16.60 & 15.60 & 50.57 & 39.78 & 34.18\\
    \hline
    \end{tabular}
    \vspace{-2.0ex}
\end{minipage}
\hspace{1mm}
\begin{minipage}{.5\linewidth}
    \caption{Results on the KITTI test set.}
    \vspace{-1.0ex}
    \centering
    \begin{tabular}{c|c|c|c}
    \hline
     \multirow{2}*{Method} & \multicolumn{3}{c}{AP$_{3D}$ IOU$\ge 0.7$}\\
    \cline{2-4}
    ~ & Easy & Mod. & Hard\\
    \hline
    MonoDIS~\cite{MonoDIS} & 10.37 & 7.94 & 6.40\\
    M3D-RPN~\cite{M3D-RPN} & 14.76 & 9.71 & 7.42\\
    MonoPair~\cite{MonoPair} & 13.04 & 9.99 & 8.65\\
    MoVi-3D~\cite{MoVi-3D} & 15.19 & 10.90 & 9.26\\
    RTM3D~\cite{RTM3D} & 14.41 & 10.34 & 8.77\\
    PGD (Ours) & \textbf{19.05} & \textbf{11.76} & \textbf{9.39}\\
    \hline
    \end{tabular}
    \vspace{-2.0ex}
    \label{tab: KITTI_test}
\end{minipage}
\end{table}

\begin{table}
\tiny
\begin{minipage}{.5\linewidth}
    \centering
    \caption{Ablation study on nuScenes.}
    \vspace{-1.0ex}
    \label{tab: nus_ablation}
	\begin{tabular}{c|c|c|c|c|c|c}
	\hline
	Methods & mAP & mATE & mASE & mAOE & mAAE & NDS\\
	\hline
	FCOS3D~\cite{FCOS3D} & 0.319 & 0.743 & 0.265 & 0.543 & 0.155 & 0.389\\
	+Local cons. & 0.325 & 0.721 & 0.266 & 0.546 & 0.164 & 0.393\\
	+Prob. depth & 0.339 & 0.716 & 0.265 & 0.511 & 0.163 & 0.404\\
	+Depth prop. & 0.348 & 0.701 & 0.268 & 0.452 & 0.166 & 0.415\\
	\hline
	\end{tabular}
	\vspace{-6ex}
\end{minipage}
\hspace{1mm}
\begin{minipage}{.48\linewidth}
    \centering
    \caption{Ablation study for the depth score.}
    \vspace{-1.0ex}
    \label{tab: ablation_ds}
    \begin{tabular}{c|c|c|c}
    \hline
     \multirow{2}*{Method} & \multicolumn{3}{c}{AP$_{3D}$ IOU$\ge 0.7$}\\
    \cline{2-4}
    ~ & Easy & Mod. & Hard\\
    \hline
     Top-2 score & 20.58 & 16.30 & 14.99\\
     Norm. Entropy & 20.15 & 15.89 & 14.67\\
     1 - Std. & 20.68 & 16.26 & 14.94\\
    \hline
    \end{tabular}
    \vspace{-6ex}
\end{minipage}
\end{table}
\begin{table}[thb]
\tiny
\begin{minipage}{.48\linewidth}
    \centering
    \vspace{-1ex}
    \caption{Ablation study for probabilistic depth.}
    \vspace{-2ex}
    \label{tab: ablation_prob}
    \begin{tabular}{c|c|c||c|c|c}
    \hline
     prop. branch & w/ direct & depth score & Easy & Mod. & Hard\\
     \hline
    \checkmark & & & 14.53 & 11.93 & 10.70\\
    \checkmark & \checkmark & & 16.58 & 13.82 & 13.49\\
    \checkmark & \checkmark & \checkmark & 19.10 & 16.04 & 14.83\\
    \hline
    \end{tabular}
    \vspace{-3.5ex}
\end{minipage}
\hspace{1mm}
\begin{minipage}{.48\linewidth}
    \centering
    \vspace{-2ex}
    \caption{Ablation study for geometric depth.}
    \vspace{-1.0ex}
    \label{tab: ablation_geo}
    \begin{tabular}{c|c|c||c|c|c}
    \hline
     fusion & edge gating & cut off grad. & Easy & Mod. & Hard\\
     \hline
    \checkmark & & & 18.54 & 15.44 & 13.94\\
    \checkmark & \checkmark & & 19.50 & 15.87 & 14.19\\
    \checkmark & \checkmark & \checkmark & 21.36 & 16.60 & 15.60\\
    \hline
    \end{tabular}
    \vspace{-3.5ex}
\end{minipage}
\end{table}
\subsection{Ablation Studies}
\vspace{-0.8ex}
Finally, we conduct ablation studies to validate the efficacy of our proposed key components on KITTI (Tab.~\ref{tab: KITTI_ablation}) and nuScenes (Tab.~\ref{tab: nus_ablation}). We can observe that local constraints can basically enhance the baseline, and our probabilistic and geometric depth further boost the performance significantly, especially in terms of mAP and translation error (mATE). Tab.~\ref{tab: ablation_prob} and Tab.~\ref{tab: ablation_geo} show more details of two core components for improving depth estimation with the metrics average precision under IOU$\ge$0.7. It can be seen that combining the probabilistic representation (prop. branch in Tab.~\ref{tab: ablation_prob}) with direct regression (w/ direct) and leveraging the depth score in the inference (depth score) can finally make the most of this design. For geometric depth, the basic fusion with local estimation can not bring the desirable gain. Improving the propagation graph via edge pruning and gating (edge gating) and cutting off the unexpected gradients propagation (cut off grad.) can help remove possible noises and prompt the learning more focused on the final integration, thus making the overall scheme much more effective.
\vspace{-0.5ex}
As for alternative implementations, we compare feasible methods of computing the depth score from the probabilistic distribution (Tab.~\ref{tab: ablation_ds}). Compared to other more complicated ways, normalized entropy and standard deviation, our exploited top-2 score can achieve decent results with better efficiency. See more results about different depth division methods and detailed analyses for these two datasets from other perspectives like the Precision-Recall curve in the appendix.
% !TEX root = ./arxiv.tex

\section{Conclusion}
This paper targets the key challenge lying in monocular 3D object detection, instance depth estimation.
Started from a basic adapted 3D detector, we firstly make in-depth oracle analyses. We surprisingly find that depth estimation is the dominating bottleneck for current 3D detection, especially in terms of localization.
To tackle the discovered challenge, we propose a novel approach, Probabilistic and Geometric Depth (PGD), which leverages the geometric relationship in perspective to construct a graph connecting instance estimations with uncertainty and thus predicts depths more accurately. The efficacy of this solution is demonstrated on both KITTI and large-scale nuScenes datasets. In the future, we will further extend the geometric depth scheme to more general cases by relaxing the "ground" assumption via 2D height regression or ground normal estimation, and validate this pipeline on other 2D detectors. How to better leverage temporal geometry information to address the difficulty of instance depth estimation is also a promising direction worthy of further exploration.
\label{sec:conclusion}

\clearpage
% The acknowledgments are automatically included only in the final version of the paper.
\acknowledgments{This work is supported in part by Centre for Perceptual and Interactive Intelligence Limited, in part by the GRF through the Research Grants Council of Hong Kong under Grants (Nos. 14208417, 14207319 and 14203518) and ITS/431/18FX, in part by CUHK Strategic Fund and CUHK Agreement TS1712093, in part by the Shanghai Committee of Science and Technology, China (Grant No. 20DZ1100800).}

\bibliography{ref}  % .bib

% !TEX root = ./arxiv.tex

\clearpage
\begin{center}
    \Large
    \textbf{Appendix}
\end{center}

\begin{figure}[thb]
\begin{center}
\includegraphics[width=1.0\linewidth]{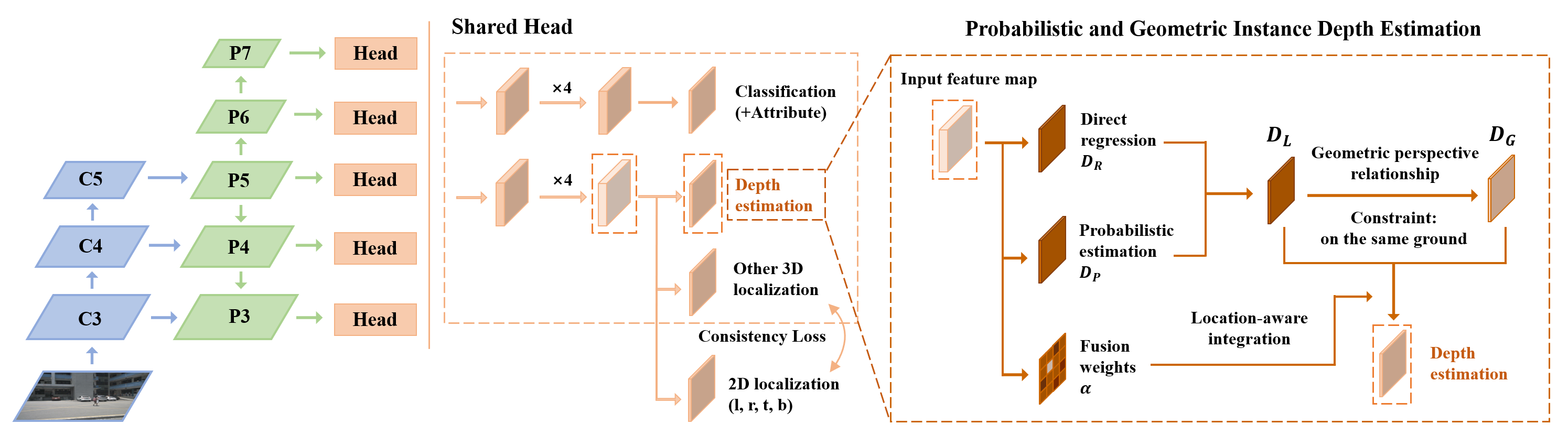}
\end{center}
   \vspace{-2ex}
   \caption{An overview of our framework (Figure 3 in the main paper).}
\label{fig: apdx_overview}
\vspace{-1ex}
\end{figure}

%===============================================================================
\setcounter{section}{0}
\section{Implementation Details}
This section first presents the adopted local geometric constraints between 2D and projected 3D bounding boxes in the enhanced baseline. Subsequently, we will elaborate on the details of training loss and inference
procedure.
\subsection{Local Geometric Constraints}
Our baseline FCOS3D~\cite{FCOS3D} only stiffly adjusts the output of networks to fit the requirements of 3D detection. There is no relationship or constraints between these predicted attributes, making this network hard to train, especially when the data is limited. Considering our detector can achieve 90\% accuracy on 2D vehicle detection, we add 2D localization into our targets and use it to regularize 3D outputs. Actually, this closed-loop and self-supervised approach is also consistent with what humans do in the annotation procedure~\cite{FLAVA}. In practice, as shown in Fig.~\ref{fig: apdx_overview}, we add a consistency loss (GIoU loss) between our estimated 2D boxes and the exterior 2D boxes of 3D predictions to enhance our baseline, which is particularly important on the small KITTI dataset. Note that due to the difficulty of regressing accurate depth, we use the ground truth depth for deriving the 3D bounding boxes when computing the consistency loss.

\begin{figure*}
\begin{center}
\includegraphics[width=0.9\linewidth]{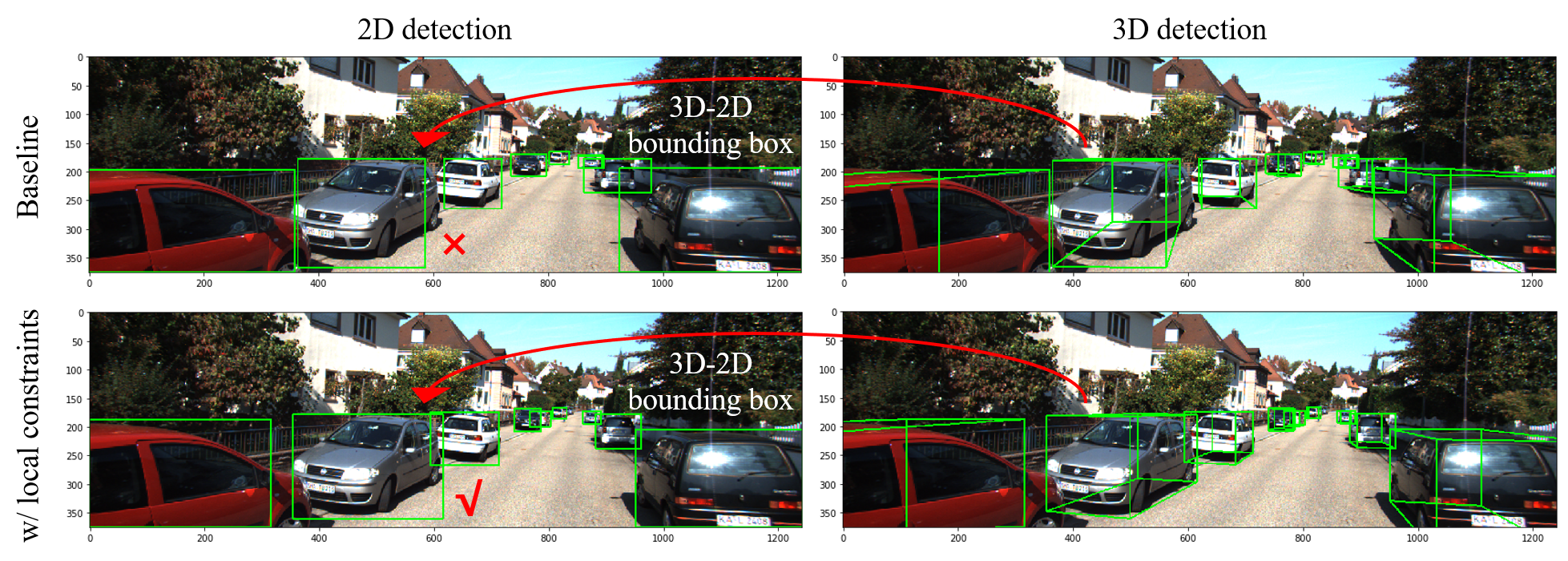}
\end{center}
   \vspace{-2ex}
   \caption{The top line shows that it is easy to validate the accuracy of 3D predictions according to its exterior 2D bounding box. So we add the 2D localization into our targets and use the relatively reliable 2D boxes to regularize 3D predictions. This results in significant improvement as shown by the bottom line.}
\label{fig: local_geo}
\vspace{-2ex}
\end{figure*}

Here we provide an example to show the intuition behind this design. Typically when the data is limited, it is hard for the network to direct regress different 3D targets (offset, depth, orientation, \emph{etc.}) independently. For example, in Fig.~\ref{fig: local_geo}, the orientation of nearby large objects predicted by our baseline can be very inaccurate (the top line in the figure) even though it can be easily rectified with simple verification. So we add the more reliable 2D localization into our targets to regularize our 3D predictions. It turns out that the simple local constraint could alleviate this problem in the learning procedure while does not introduce extra computational costs to inference. The improved results after adding this constraint can be seen in Fig.~\ref{fig: local_geo} (the bottom line).

\subsection{Loss}
\noindent\textbf{Overall Loss Design}\quad We basically follow the loss design of FCOS3D except our proposed consistency loss and the adjustments for different datasets.

To have a brief review, firstly, we use the focal loss~\cite{RetinaNet} as the object classification loss:
\begin{equation}
    \centering
    L_{cls} = -\alpha(1-p)^\gamma logp
\end{equation}
where $p$ is the class probability of a predicted box, and we follow the common settings, $\alpha = 0.25$ and $\gamma = 2$. For attribute classification on nuScenes, we use a simple softmax classification loss, denoted as $L_{attr}$.

For regression branch, we use the smooth L1 loss for each regression target except centerness:
\begin{equation}
    \label{eqn: loc_loss}
    \centering
    L_{loc} = \sum_{b\in (\Delta x, \Delta y, d, w, l, h, \theta, v_x, v_y)} SmoothL1(\Delta b)
\end{equation}
The weights of $\Delta x, \Delta y, d, w, l, h, \theta$ error are 1 and the weights of $v_x, v_y$ on nuScenes are 0.05. We use the softmax classification loss and binary cross entropy (BCE) loss for direction classification and centerness regression, denoted as $L_{dir}$ and $L_{ct}$ respectively. For local geometric constraints, denote our predicted 2D boxes as $\bm{B_{2D}}$, the minimum exterior 2D boxes of projected 3D boxes as $\bm{B_{proj}}$, then the consistency loss is:
\begin{equation}
    \label{eqn: cons_loss}
    \centering
    L_{geo} = GIoU(\bm{B_{2D}}, \bm{B_{proj}})
\end{equation}
Finally, the total loss is:
\begin{equation}
    \centering
    L = \frac{1}{N_{pos}}(\beta_{cls}L_{cls}+\beta_{attr}L_{attr}+\beta_{loc}L_{loc}+\beta_{dir}L_{dir}+\beta_{ct}L_{ct}+\beta_{geo}L_{geo})
\end{equation}
$N_{pos}$ is the number of positive predictions and $\beta_{cls} = \beta_{attr} = \beta_{loc} = \beta_{dir} = \beta_{ct} = \beta_{geo} = 1$. Note that the attribute loss $L_{attr}$ and velocity loss in the $L_{loc}$ are only required in the nuScenes experiments.

\noindent\textbf{Specific Loss Designs for KITTI experiments}\quad Because the KITTI dataset has relatively limited samples and much more strict metrics, we adopt two specific loss designs for training the networks. First, we add an auxiliary key-points loss to enhance the local geometric consistency further. Denote the 2D offsets of eight key-points (eight corners of a 3D bounding box) relative to a foreground point as $\bm{k}\in\mathbb{R}^{1\times 16}$, and then we take these offsets as 16 additional dimensions of $b$ in Eqn.~\ref{eqn: loc_loss} and set their weights to 0.2. To make the FPN-based learning stable, we normalize these offsets just as we normalize those offsets to four sides of a 2D box.

In addition, we use a much stronger uncertainty formulation for this multi-task learning problem as presented in~\cite{MultiTask}. Specifically, referring to its formulation of maximum likelihood and homoscedastic uncertainty, we formulate the depth loss as:
\vspace{-1.0ex}
\begin{equation}
    L_{depth} = \frac{L_1(\hat{D}, D)}{2\sigma^2} + log\sigma
    \vspace{-1.0ex}
    \label{eqn: homo}
\end{equation}
Here $\hat{D}$ and $D$ are the targets and predictions of depth, $L_1$ represents the original smooth L1 loss with $\delta=3.0$ and $\sigma$ is the variable for uncertainty. In practice, to make the learning easier, we train the network to predict the log variance $s=log\sigma^2$ only for depth estimation, which is more numerically stable than directly predicting the variance. Correspondingly, $exp(-s)$ serves as the weight of depth loss. In this way, the depth loss will be adaptively weighted relative to other regression losses. Additionally, the uncertainty $exp(-s)$ can also be used as another confidence score to be multiplied when inference, such that predictions with more accurate depths will have particularly higher scores. Note that this strong uncertainty indicator can only bring a significant gain on KITTI experiments while seriously hurting the general performance as evaluated on the nuScenes dataset.  

\noindent\textbf{Alternative Depth Loss Designs}\quad Considering we have several intermediate depth predictions, such as $D_R$, $D_P$ and $D_L$ in Fig.~\ref{fig: apdx_overview}, a natural idea is to add intermediate supervisions for these predictions to guarantee that each branch can learn meaningful information. So we further defined several depth L1 losses for these predictions and tried to replace the original depth loss in the $L_{loc}$ with their weighted summation. It turns out that although this approach can make the training procedure more stable, it does not bring any performance gain. We also find that the framework never overfits to only relying on one kind of estimation even with only supervision for the final prediction, as to be shown in Sec.~\ref{sec: quant_depth}. It indicates that these predictions and components indeed work together from complementary aspects. 

\subsection{Inference}
The inference procedure is to forward the input image through the framework and obtain bounding boxes with their class scores, attribute scores (if necessary) and centerness predictions. We multiply the class score, the predicted centerness and the depth confidence score as the overall confidence for each prediction and conduct rotated Non-Maximum Suppression (NMS) in the bird view as most 3D detectors to get the final results.

%------------------------------------
\section{Explanation of Oracle Analyses}
In this section, we will explain more about our empirical analysis, from the specific settings to more details in the results.

\subsection{Reason for Replacing Dense Predictions}
First, we would like to emphasize one detail in our analysis, \emph{i.e.}, we replace the dense predictions from the direct output of detection head with oracles to purely observe the problems of our networks. In comparison, other alternatives exist, such as replacing the decoded dense output or predictions after post-processing, which can not reveal some entangling problem lying in the formulation. One example to show the difference between these two implementations is that we replace the offset with corresponding ground truth while the latter approach replaces the decoded $X, Y$ in the 3D space with targets.

\subsection{Comparison of Different Metrics}
As mentioned in the main paper, KITTI and nuScenes adopt different evaluation metrics. The former is relatively strict and the latter is more comprehensive. Specifically, for mAP of these two datasets, we regard predictions with 3D IoU larger than a threshold (0.7 or 0.5) as positive samples on KITTI while define the match by 2D center distance $d_{2D}$ in the bird eye view on nuScenes. The latter is a simpler criterion as it decouples the detection from object size and orientation. Therefore, we only plot points with category/location related oracles (classification, depth and offset) in the mAP analysis on nuScenes (Fig.~\ref{fig: apdx_oracle}). In addition, to be more specific, mAP is computed over several different matching thresholds, $\mathbb{D} = \{0.5, 1, 2, 4\}$ meters, and all categories $\mathbb{C}$ on nuScenes:\\
\begin{equation}
    \centering
    mAP = \frac{1}{|\mathbb{C}||\mathbb{D}|}\sum_{c\in\mathbb{C}}\sum_{d_{2D}\in\mathbb{D}}AP_{c,d_{2D}}
\end{equation}
Then we can see that it will also consider predictions with relatively inaccurate locations (like objects with the distance error larger than 2 meters but smaller than 4 meters). This difference is especially notable when discussing the improvements from depth score, which will be detailed in Sec.~\ref{sec: PR_curve}.

Finally we basically describe how the NuScenes Detection Score (NDS) is calculated. To begin with, we first define that predictions with center distance from the matching ground truth $d_{2D} \le 2m$ will be considered as true positives (TP) and thus introduce 5 True Positive metrics, Average Translation Error (ATE), Average Scale Error (ASE), Average Orientation Error (AOE), Average Velocity Error (AVE) and Average Attribute Error (AAE). Given these metrics, we compute the mean TP metric (mTP) over all categories:\\
\begin{equation}
    \centering
    mTP = \frac{1}{|\mathbb{C}|}\sum_{c\in\mathbb{C}}TP_c
\end{equation} 
Then the NDS is calculated as follows:\\
\begin{equation}
    \centering
    NDS = \frac{1}{10}[5mAP+\sum_{mTP\in\mathbb{TP}}(1-min(1, mTP))]
\end{equation}
Therefore, NDS is a combination of several decoupled metrics and could reflect the performance of 3D detectors from another perspective. See more details about the intermediate computation in its original paper~\cite{nuScenes}.

\subsection{Detailed Explanations and Conclusions}
\begin{figure}
\begin{center}
\includegraphics[width=1.0\linewidth]{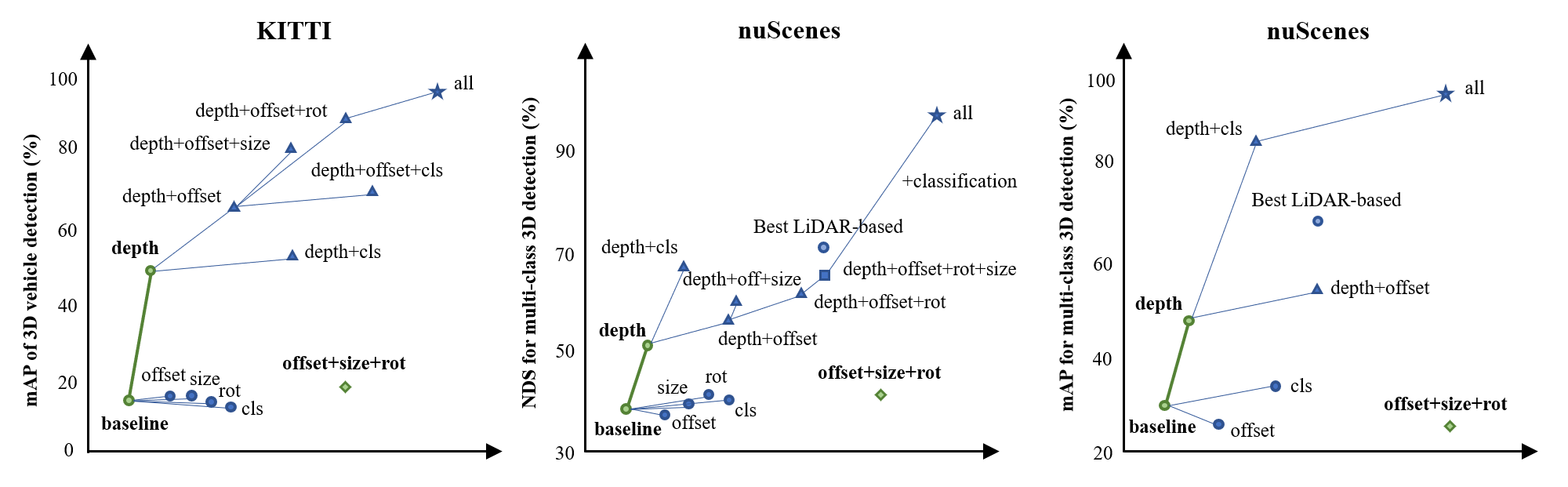}
\end{center}
   \vspace{-2ex}
   \caption{Oracle analyses with different datasets and metrics (Figure 2 in the main paper). From left to right: 3D IoU based mAP on KITTI, NuScenes Detection Score (NDS) and distance-based mAP on nuScenes. We replace our predictions with ground truth values step by step and observe the performance improvements. It can be seen that an accurate depth can bring significant performance improvement (green lines), and only with accurate depth can the improvements brought by other oracles be realized.}
\label{fig: apdx_oracle}
\vspace{-2ex}
\end{figure}

Due to the space limitation in the main paper, we do not discuss much about the results shown in Fig.~\ref{fig: apdx_oracle}. Next, we will analyze it in detail and summarize a series of important conclusions.

\noindent\textbf{Basic Observations}\quad As shown in Fig.~\ref{fig: apdx_oracle}, we replace the predicted attributes with their ground truth values step by step and observe the performance improvements. We can see that:

\noindent 1. With only one oracle (circle dots), only depth can bring a considerable improvement (green lines). It shows that with current depth estimation, other predicted attributes do not drag down the performance, while with other predictions, the current accuracy of depth estimation is far not enough.

\noindent 2. With accurate depth, other oracles (triangle dots in the figures) could bring the expected performance gains. While with current depth estimation, even all the other predictions are accurate (green rhombus dots), the results are always disappointing, even almost like the baseline.

\noindent 3. Although KITTI and nuScenes are different in terms of category variety and metrics, the trend of these curves is the same. The difference is reflected in the importance of localization and classification oracles. Localization is more important on the KITTI, which has less category variety and more strict metrics. Classification is another important factor apart from depth on nuScenes, \emph{e.g.}, our monocular predictions with location oracle is still not better than the best LiDAR-based methods. In contrast, with an accurate depth and classification map, the performance is almost ideal.

From these observations, we can conclude that the inaccurate depth blocks \emph{all} the other sub-task predictions from improving the overall detection performance. Hence, as mentioned in the main paper, the current monocular 3D detection, especially 3D localization, can be actually reduced to the dominating instance depth estimation problem.

\noindent\textbf{Comparison with Best LiDAR-Based Methods}\quad There is an interesting phenomenon not much related to depth estimation in the above analysis, \emph{i.e.}, the comparison with best LiDAR-based methods on nuScenes. We can see that classification is particularly important on nuScenes, and our monocular predictions with location oracles are still not better than the state-of-the-art LiDAR-based methods. This result is a little dataset-specific. We conjecture it is because the classification for ten categories on nuScenes is relatively hard, or the annotation is mainly conducted in the point clouds, leading to missing objects in the images.
%------------------------------------

\section{Supplementary Experimental Results}
In this section, we will show more experimental results to help further understand our approach. First, we will provide toy examples to explain and validate our derived pairwise perspective relationship in the depth propagation. Subsequently, we make more detailed analyses in quantitative and qualitative ways to reveal the working mechanism and effect of our method.

\begin{figure}
\begin{center}
\includegraphics[width=1.0\linewidth]{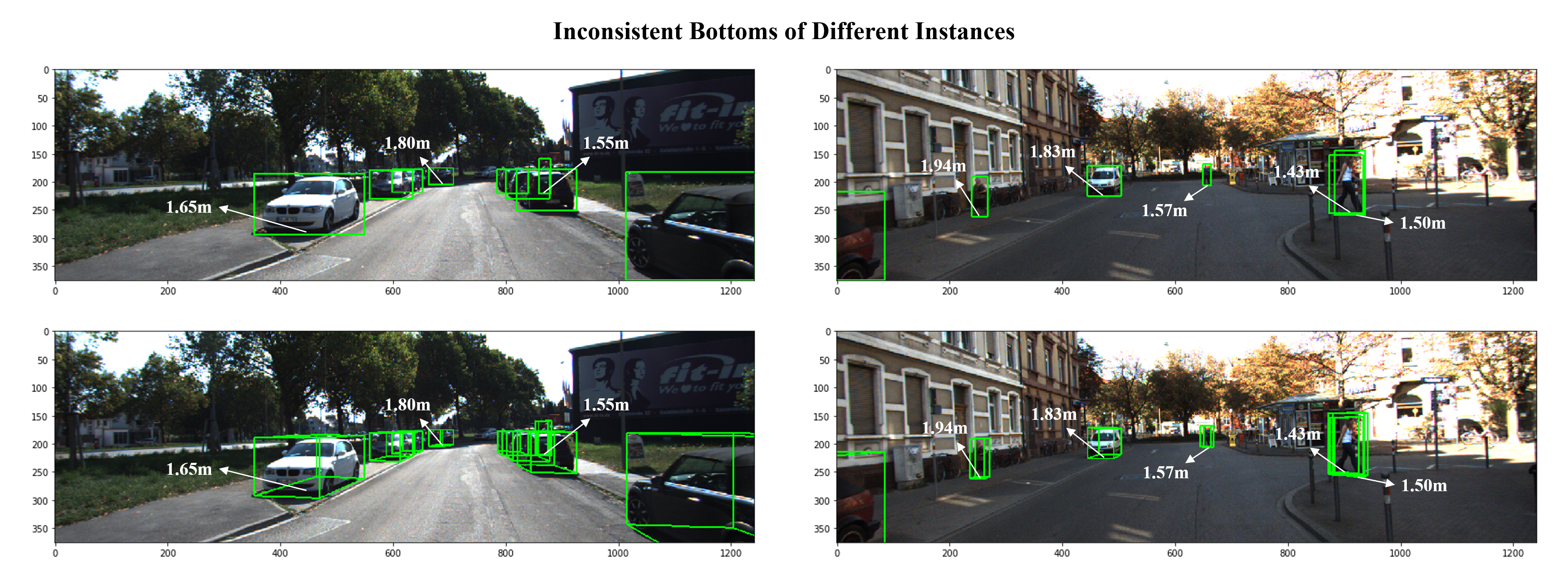}
\end{center}
   \vspace{-3ex}
   \caption{Inconsistent bottoms of different instances. Although all the objects in an image share similar heights for bottoms most of the time, corner cases still exist. Here we mark the heights of bottoms in the camera coordinates (down is the positive direction). This problem can be caused by the actual topography, \emph{e.g.}, pedestrians are on the step. It can also be caused by annotation noises, especially for different categories and distant objects. This observation is the foundation of our proposed edge pruning/gating scheme in the depth propagation.}
\label{fig: toy_exp}
\vspace{-1.5ex}
\end{figure}
\begin{figure}
\begin{center}
\includegraphics[width=1.0\linewidth]{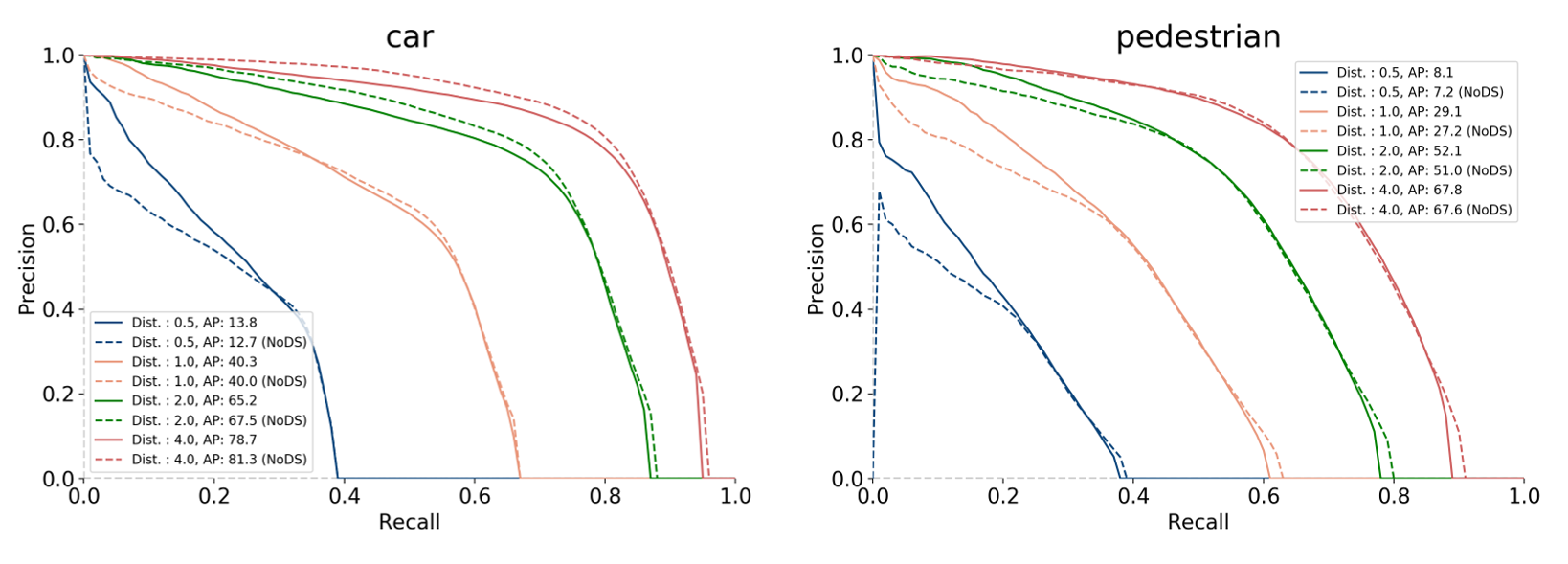}
\end{center}
   \vspace{-4ex}
   \caption{Comparison of PR curves for models with (solid line) and without (dotted line) depth score. The depth score encourages predictions with accurate depth while suppresses those with inaccurate depth, which results in higher precision under low recall and strict matching thresholds while lower precision under high recall. This problem is more notable for large objects like cars.}
   \vspace{-2.5ex}
\label{fig: PR_curve}
\end{figure}

\subsection{Basic Validation of Depth Propagation}
As shown in Fig.~\ref{fig: toy_exp}, we provide two samples with many objects in one image. We first have a brief review of the perspective relationship derived in the main paper. Given two objects 1 and 2, the relationship between their centers strictly satisfies:
\begin{equation}
    d_2 = \frac{v_1}{v_2}d_1 + \frac{f}{v_2}(y_2-y_1)
\end{equation}
Considering two objects share the same ground (bottom height), we can get the approximate relationship as follows:
\begin{equation}
    d_2 = \frac{v_1}{v_2}d_1 + \frac{f}{2v_2}(h^{3D}_1-h^{3D}_2)
    \label{eqn: depth_prop}
\end{equation}
where $d$ denotes the depth, $v$ denotes the distance between the projected 2D object center and the horizon line in the image, $y$ is the 3D height of object center and $h^{3D}$ is the height of the 3D bounding box. Taking the left sample in Fig.~\ref{fig: toy_exp} as an example, the depths of the 8 cars are $\{5.23, 11.80, 16.50, 22.05, 23.64, 28.53, 29.07, 42.85\}$. With our derived relationship, we can estimate them with only the first 2 accurate depths: $\{5.23, 11.74, 16.78, 22.92, 21.13, 26.59, 25.78, 36.51\}$. We can see that similar to the general case of depth estimation, our propagation mechanism also yields more notable errors for distant objects, which has been analyzed in the main paper (The effect of $\delta$ over $\Delta d$ will be enlarged as the $v_2$ decreases.)

Next, we can further observe the inconsistent bottoms problem shown in Fig.~\ref{fig: toy_exp}. We mark some representative instances in the figure. It can be seen that it is sometimes caused by the actual topography, like pedestrians and cars in the second sample. Nevertheless, the noise only exists between objects far away from each other most of the time. We conjecture this is related to the annotation pipeline, \emph{e.g.}, we tend to make use of nearby annotations when the information for labeling the current instance is inadequate. Alternatively, sometimes it is just because the LiDAR only sweeps the top part of the distant objects such that the annotator can not determine its bottom accurately.

In conclusion, although the ground constraint holds most of the time, it is still important to design mechanisms to avoid these possible noises and incorporate the geometric depth adaptively, such as the edge pruning/gating scheme and location-aware integration in the main paper.

\vspace{-0.5ex}
\subsection{Quantitative Analysis}
\vspace{-0.5ex}
\noindent\textbf{Difference Between Datasets}\label{sec: PR_curve}\quad
Here we mainly show the observation in the ablation study to explain the different effects from the same component on these two datasets. We take the depth score as an example. First, Tab. 7 in the main paper has shown the especially important role of depth score on the KITTI. However, it does not contribute much to the improvements on nuScenes. Specifically, it only brings about 0.3\% increase on NDS by reducing the mATE instead of boosting the mAP. To figure out the reason, we take a closer look at the performance from the Precision-Recall (PR) curve. As shown in Fig.~\ref{fig: PR_curve}, we can see that the depth score (solid line) significantly improves the precision under low recall and strict matching thresholds (like 0.5 and 1.0 meters, blue and yellow lines) while influences the performance under high recall and less strict cases (like 2.0 and 4.0 meters, green and red lines). This problem is especially notable for large objects. It reveals the effect of depth score from another perspective, \emph{i.e.}, it can overly suppress those predictions with inaccurate depth, of which we should be tolerant under some circumstances, like distant and small objects. Therefore, designing a more suitable depth score with better interval division methods or other approaches can be a direction worthy of further exploration.

\noindent\textbf{Mean AP for Multi-Class Detection on nuScenes}\quad
To present the multi-class detection results more comprehensively, we provide the mean AP results (over all the matching thresholds) for each category on nuScenes in Tab.~\ref{tab: ap_class}. We can see that our method shows the superiority especially on small (from pedestrian to barrier) and quite large objects (bus). Firstly, the better capability of handling objects with different scales should partly come from the leveraged well-developed backbone and FPN. Furthermore, our probabilistic and geometric depth also improves the accuracy of depth estimation, which is especially important for small objects.

\begin{table}\scriptsize
\caption{Average precision for each class on the nuScenes test benchmark. CV and TC are abbreviation of construction vehicle and traffic cone in the table.}
	\begin{center}
	\begin{tabular}{c|c|c|c|c|c|c|c|c|c|c|c}
	\hline
	Methods & car & truck & bus & trailer & CV & ped & motor & bicycle & TC & barrier & mAP\\
	\hline\hline
	LRM0 & 0.467 & 0.21 & 0.17 & 0.149 & 0.061 & 0.359 & 0.287 & 0.246 & 0.476 & 0.512 & 0.294\\
	\hline
	MonoDIS~\cite{MonoDIS} & 0.478 & 0.22 & 0.188 & 0.176 & 0.074 & 0.37 & 0.29 & 0.245 & 0.487 & 0.511 & 0.304\\
	\hline
	CenterNet~\cite{CenterNet} (HGLS) & 0.536 & 0.27 & 0.248 & 0.251 & 0.086 & 0.375 & 0.291 & 0.207 & 0.583 & 0.533 & 0.338\\
	\hline
	Noah CV Lab & 0.515 & 0.278 & 0.249 & 0.213 & 0.066 & 0.404 & 0.338 & 0.237 & 0.522 & 0.49 & 0.331\\
	\hline
	PGD (Ours) & 0.561 & 0.299 & 0.285 & 0.266 & 0.134 & 0.441 & 0.397 & 0.314 & 0.605 & 0.561 & \textbf{0.386}\\
	\hline
	\end{tabular}
	\end{center}
	\vspace{-5ex}
	\label{tab: ap_class}
\end{table}

\begin{figure}
\begin{minipage}{.48\linewidth}
    \centering
    \includegraphics[width=1.0\linewidth]{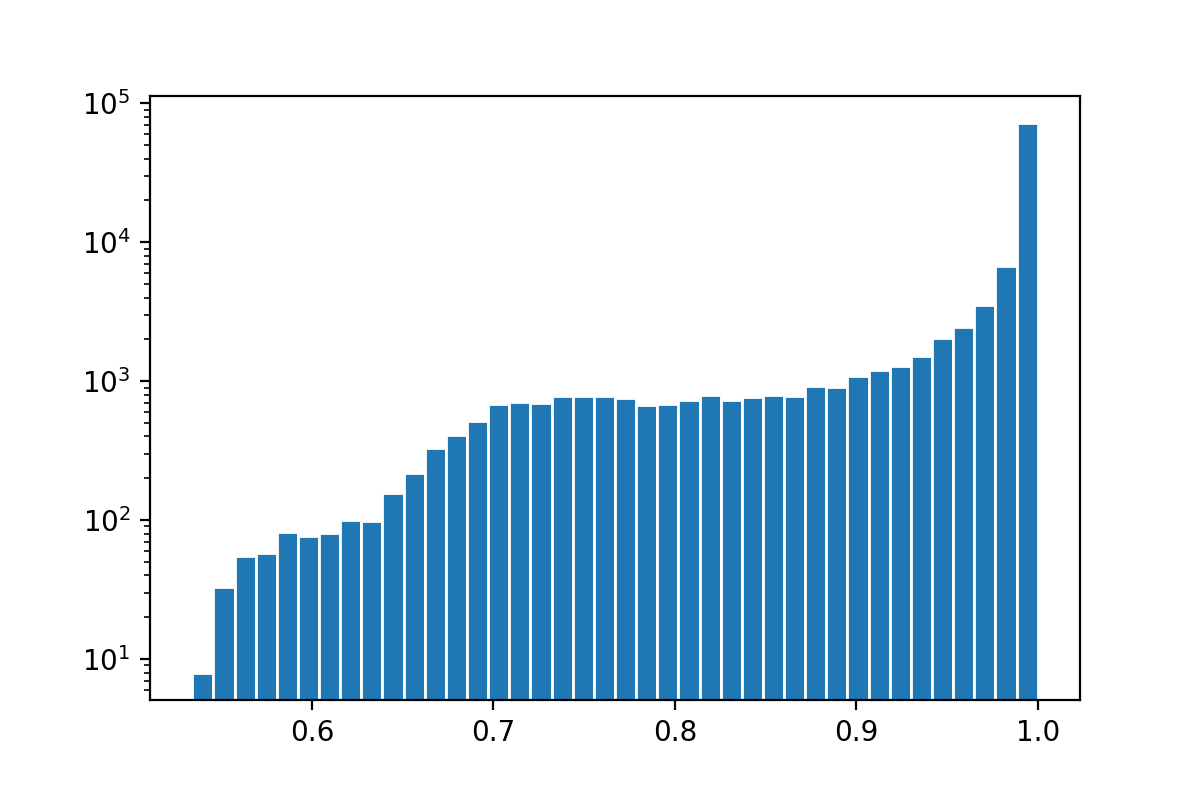}
    \vspace{-5ex}
    \caption{Distribution of weights for the final integration in our PGD module.}
    \vspace{-2ex}
    \label{fig: prop_weights}
\end{minipage}
\hspace{1mm}
\begin{minipage}{.48\linewidth}
    \centering
    \includegraphics[width=1.0\linewidth]{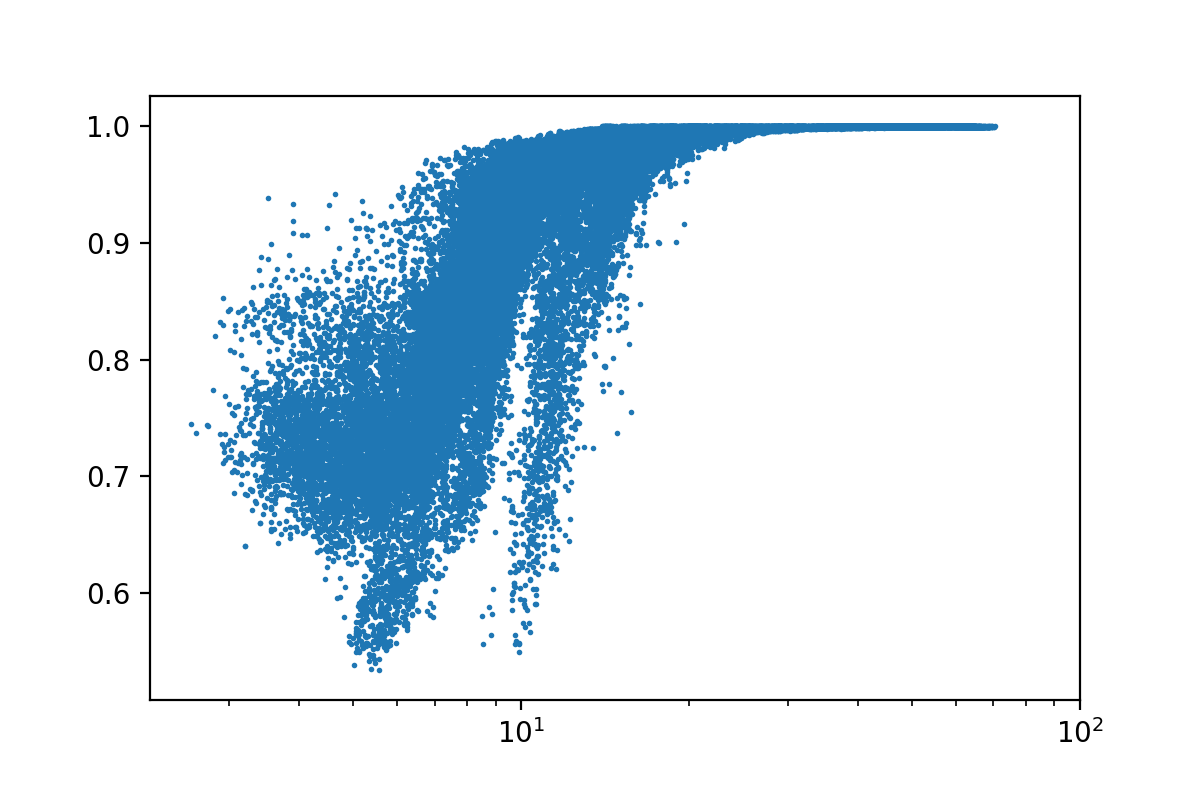}
    \vspace{-5ex}
    \caption{Scatter plot of location-aware weights with respect to depths.}
    \vspace{-2ex}
    \label{fig: weights_vs_depths}
\end{minipage}
\end{figure}

\begin{figure}
\begin{center}
\includegraphics[width=1.0\linewidth]{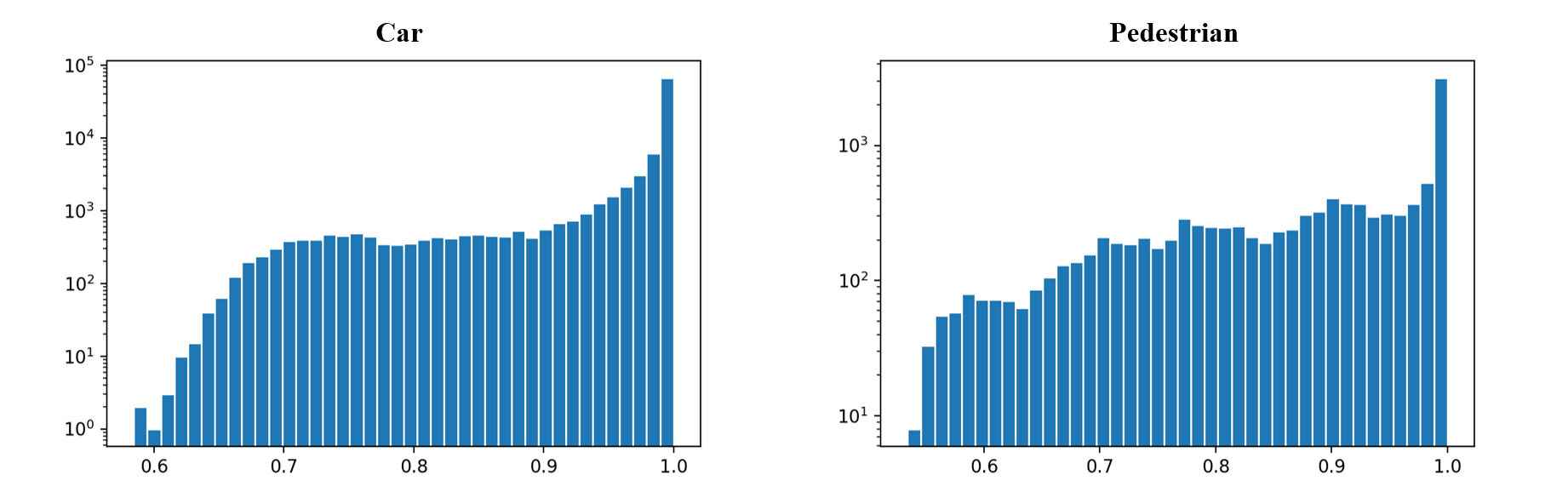}
\end{center}
    \vspace{-3ex}
    \caption{Location-aware weights of predictions from different categories.}
    \vspace{-3ex}
\label{fig: weights_cls}
\end{figure}

\noindent\textbf{Contributions of Each Depth Estimation}\label{sec: quant_depth}\quad
To understand the role of each component for depth estimation more clearly, we make statistics about the fusion weights. Firstly, for local depth estimation, we find that the direct regression accounts for about 25.6\% in the results, \emph{i.e.}, $\sigma(\lambda)$ is about 0.256. It implies that the direct regression may be responsible for regressing the residual of the probabilistic estimation, which plays an auxiliary but important role according to the ablation study in the main paper (Tab. 7). On the other hand, for final integration, we make statistics for the location-aware weights $\sigma(\alpha)$ of predictions with matching ground truths before NMS on the validation set, and plot its distribution in Fig.~\ref{fig: prop_weights} (higher value means more contribution from local estimation). We can see that although the preliminary local estimation plays a more important role in many cases, the propagated geometric depth does contribute a lot to the overall estimation. In addition, we also plot the scatter diagram of these weights with respect to the estimated depth and different categories (Fig.~\ref{fig: weights_vs_depths} and \ref{fig: weights_cls}). We can see that the geometric depth contributes more to the estimation of very nearby (can be truncated in the image) and small objects like pedestrians, which is consistent with our common sense that these two cases are relatively hard such that we need to incorporate some contextual information in the reasoning procedure.

\begin{table}[htb]
\tiny
\begin{minipage}{.5\linewidth}
    \centering
    \vspace{-1.0ex}
    \caption{Ablation study for the depth unit setting with our lightweight model on nuScenes.}
    \vspace{-1ex}
	\begin{tabular}{c|c|c|c|c|c|c}
	\hline
	$U$ (meters) & mAP & mATE & mASE & mAOE & mAAE & NDS\\
	\hline
	5 & 0.298 & 0.79 & 0.266 & 0.563 & 0.164 & 0.371\\
	10 & 0.303 & 0.775 & 0.265 & 0.548 & 0.164 & 0.376\\
	\hline
	\end{tabular}
	\label{tab: ablation_unit}
    \vspace{-2ex}
\end{minipage}
\hspace{4.5ex}
\begin{minipage}{.45\linewidth}
    \centering
    \vspace{-1.0ex}
    \caption{Ablation study for alternative depth division methods.}
    \vspace{-1.0ex}
    \begin{tabular}{c|c|c|c}
    \hline
     Methods & Easy & Mod. & Hard\\
     \hline
    Log & 9.91 & 8.68 & 7.95\\
    Linear & 18.63 & 14.49 & 13.25\\
    Uniform Log & 8.62 & 13.48 & 13.28\\
    Uniform & \textbf{19.10} & \textbf{16.04} & \textbf{14.83}\\
    \hline
    \end{tabular}
    \label{tab: ablation_div}
    \vspace{-2ex}
\end{minipage}
\end{table}

\noindent\textbf{Ablation Studies for Alternative Depth Division Methods}\quad
We also made ablation studies for alternative probabilistic depth settings, including the different settings for the depth unit $U$ and different division methods to bucket the depth value into intervals. First, Tab.~\ref{tab: ablation_unit} shows that more fine-grained division can not bring performance gains. As for the division methods, we test several alternatives as shown in Tab.~\ref{tab: ablation_div}, among which \emph{Log} and \emph{Linear} refer to the spacing-increasing discretization (SID)~\cite{DORN} and linear-increasing discretization (LID)~\cite{Center3D}, respectively. We directly take their split points and compute the depth estimation with Eqn. 1 in the main paper. In contrast, \emph{Uniform Log} means that we take the split points that are uniformly distributed in the \emph{log} space as the base to compute the depth estimation in the \emph{log} space with Eqn. 1, and then apply the exponential transformation to get the final result. We can see that although the simplicity, our adopted uniform division method achieves the best performance. Note that this ablation study is conducted with $U=10m$. There may be different conclusions if we exploit more fine-grained divisions or use classification and residual regression to implement the probabilistic depth estimation.

\begin{table}[htb]
\tiny
\begin{minipage}[htb]{.48\linewidth}
    \centering
    \vspace{-2ex}
    \caption{Ablation study for alternative distance scores in the edge gating scheme on KITTI.}
    \vspace{-1ex}
    \begin{tabular}{c|c|c|c|c|c|c}
    \hline
     \multirow{2}*{Method} & \multicolumn{3}{c|}{AP$_{3D}$ IOU$\ge 0.7$} & \multicolumn{3}{c}{AP$_{3D}$ IOU$\ge 0.5$}\\
    \cline{2-7}
    ~ & Easy & Mod. & Hard & Easy & Mod. & Hard\\
    \hline
     3D bottoms & 15.18 & 11.96 & 10.72 & 46.27 & 37.99 & 33.09\\
     3D centers & 21.04 & 16.07 & 14.89 & 47.03 & 37.58 & 32.97\\
     2D centers & 21.36 & 16.60 & 15.60 & 50.57 & 39.78 & 34.18\\
    \hline
    \end{tabular}
	\label{tab: ablation_dist}
    \vspace{-2ex}
\end{minipage}
\hspace{4.5ex}
\begin{minipage}[htb]{.48\linewidth}
    \centering
    \vspace{-2ex}
    \caption{Depth error statistics for predictions having corresponding matching ground truths.}
    \begin{tabular}{c|c|c}
    \hline
    Methods & Mean Abs. Error (m) $\downarrow$ & Mean Rel. Error $\downarrow$\\
    \hline
    FCOS3D & 0.0528 & 4.27\% \\
    PGD (Ours) & 0.0483 & 3.63\% \\
    \hdashline
    Rel. Delta & -8.5\% & -15.0\%\\
    \hline
    \end{tabular}
    \label{tab: depth_error}
    \vspace{-2ex}
\end{minipage}
\end{table}

\noindent\textbf{Ablation Studies for Geometric Depth}\quad
Recall that we select three important factors for edge pruning and gating in the depth propagation graph. We also tried other alternatives for the distance score, including the height difference between 3D bottoms, the distance of 3D centers and our adopted 2D centers (Tab.~\ref{tab: ablation_dist}). It can be observed that using the 2D centers yields the best performance. We conjecture that it is because the 3D criteria are based on the inaccurate depths such that they are less reliable than the disentangled 2D distance.

\noindent\textbf{Depth Error Analysis}\quad
We have validated the efficacy of our method in the main paper by comparing the detection performance of our method and the baseline, especially in terms of the improved mean average precision (mAP) and the mean translation error (mATE). Here we further prove its effectiveness with the depth error analysis. We make depth error statistics for the predictions (before NMS) which have corresponding ground truths on the KITTI validation set (Tab.~\ref{tab: depth_error}). We can observe that our method significantly reduces the mean error of depth estimation, both on the absolute error and relative error ((Abs. and Rel. in Tab.~\ref{tab: depth_error}).

\subsection{Qualitative Analysis}
\begin{figure*}
\begin{center}
\includegraphics[width=1.0\linewidth]{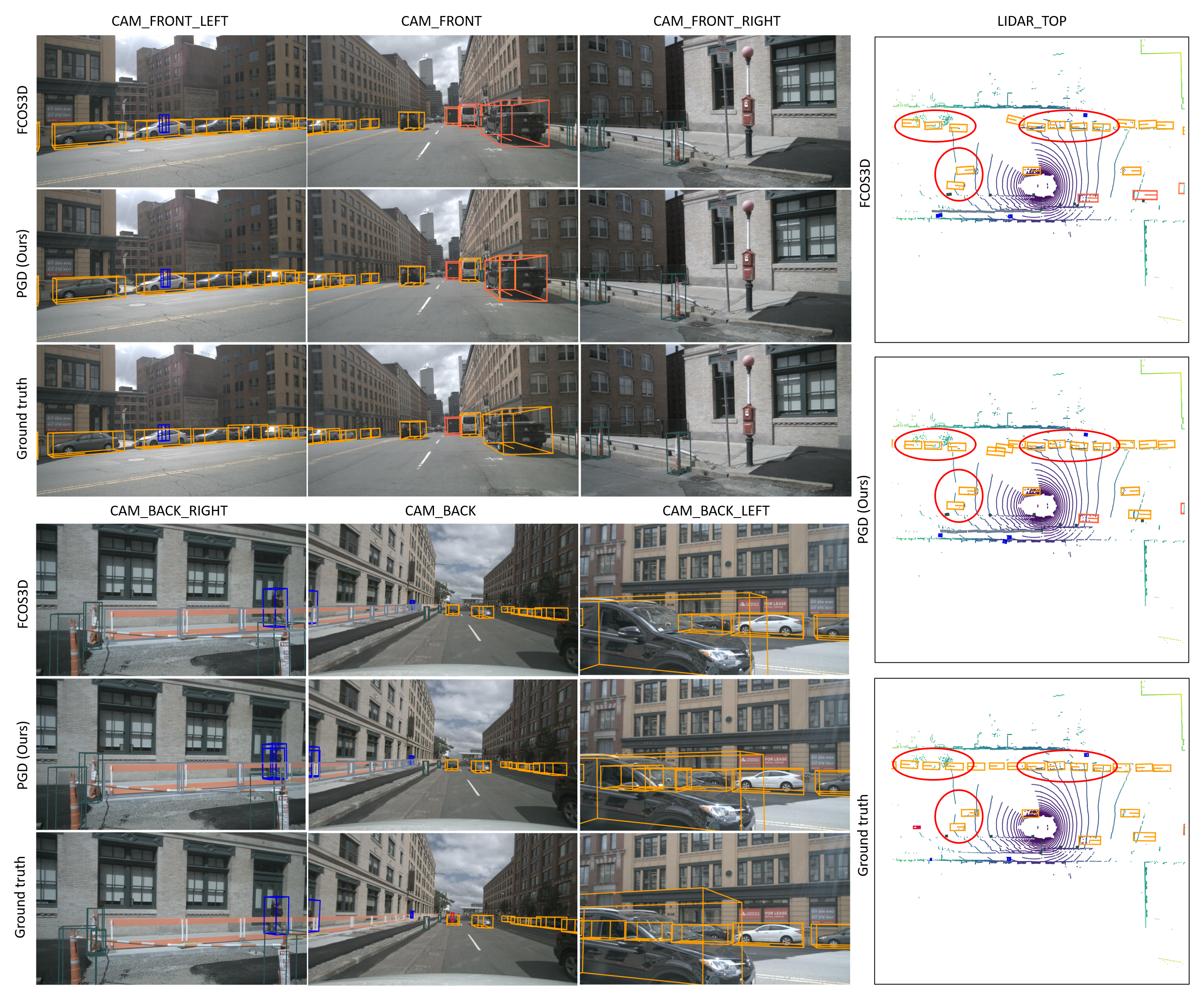}
\end{center}
   \vspace{-2ex}
   \caption{Qualitative analysis of detection results. 3D bounding box predictions are projected onto images from six different views and bird-view, respectively. Boxes from different categories are marked with different colors. We can see that the detection results of FCOS3D and PGD are both reasonable. However, from the bird-eye-view, the depth accuracy is remarkably improved by our method, especially for those objects marked with red circles.}
   \vspace{-2.5ex}
\label{fig: qualitative}
\end{figure*}

Then we show some qualitative results on nuScenes in Fig.~\ref{fig: qualitative} by drawing the predicted 3D bounding boxes in the six-view images and the top-view point clouds. We compare the results predicted by our model and the baseline FCOS3D to demonstrate the improvements in terms of depth estimation intuitively.
We can see that from the perspective of images, both detection results are appealing, especially for some small objects that are not labeled. For example, the barriers in the rear right camera are not labeled but detected by these two models.
However, from the bird-eye-view, the depth accuracy of the two methods is notably different, especially for those objects marked with red circles: The accuracy is significantly improved by our proposed method. It is also in line with the quantitative results (the mATE is reduced remarkably) and further validates the efficacy of our method.

\label{sec:appendix}

\end{document}